\documentclass[10pt,twocolumn,letterpaper]{article}

\usepackage[pagenumbers]{cvpr}

\usepackage[dvipsnames]{xcolor}
\usepackage{colortbl}
\usepackage{tabularx}
\usepackage{multicol}
\usepackage{multirow}

\usepackage{graphicx}
\usepackage{amsmath}
\usepackage{amssymb}
\usepackage{pifont}
\usepackage{booktabs}

\usepackage{titletoc}

\definecolor{robo_blue}{RGB}{99, 113, 250}
\definecolor{robo_red}{RGB}{240, 99, 75}
\definecolor{robo_green}{RGB}{0, 180, 139}

\definecolor{cvprblue}{rgb}{0.21,0.49,0.74}
\usepackage[pagebackref=true,breaklinks,colorlinks,citecolor=cvprblue,linkcolor=cvprblue]{hyperref}

\newcommand{\colorsquare}[1]{{\color{#1}$\blacksquare$}\hspace{-7.78pt}$\square$}

\usepackage[accsupp]{axessibility} 

\def\confName{CVPR}
\def\confYear{2024}

\title{OpenESS: Event-based Semantic Scene Understanding with Open Vocabularies}

\author{Lingdong Kong$^{1,2}$ \quad Youquan Liu$^{3}$ \quad Lai Xing Ng$^{4,5}$ \quad Benoit R. Cottereau$^{5,6}$ \quad Wei Tsang Ooi$^{1,5}$
\\
$^1$National University of Singapore \quad $^2$CNRS@CREATE \quad $^3$Hochschule Bremerhaven
\\
$^4$Institute for Infocomm Research, A*STAR \quad $^5$IPAL, CNRS IRL 2955, Singapore
\\
$^6$CerCo, CNRS UMR 5549, Université Toulouse III
\\
\url{https://github.com/ldkong1205/OpenESS}
}

\begin{document}

\twocolumn[{%
\renewcommand\twocolumn[1][]{#1}%
\maketitle
\begin{center}
    \centering
    \vspace{-8pt}
    \captionsetup{type=figure}
    \includegraphics[width=\textwidth]{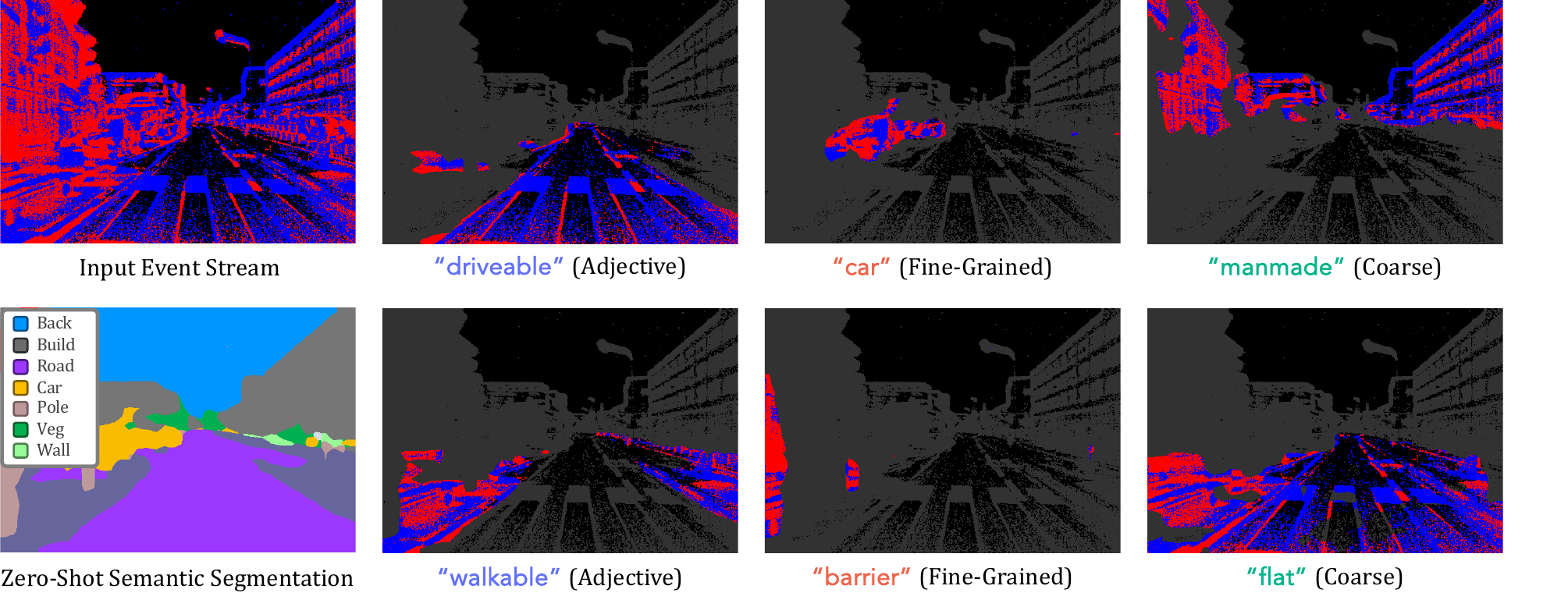}
    \vspace{-13.5pt}
    \captionof{figure}{\textbf{Open-vocabulary event-based semantic segmentation (OpenESS)}. Our framework is capable of performing zero-shot semantic segmentation of event data streams with open vocabularies. Given raw events and text prompts as inputs, OpenESS outputs semantically coherent open-world predictions across various \textcolor{robo_blue}{adjective}, \textcolor{robo_red}{fine-grained}, and \textcolor{robo_green}{coarse} categories. The last three columns show the language-guided attention maps where regions of a high similarity score to the given text prompts are highlighted. Best viewed in colors.}
    \label{fig:teaser}
    \vspace{6.36pt}
\end{center}
}]

\begin{abstract}
    \vspace{-0.125cm}
   Event-based semantic segmentation (ESS) is a fundamental yet challenging task for event camera sensing. The difficulties in interpreting and annotating event data limit its scalability. While domain adaptation from images to event data can help to mitigate this issue, there exist data representational differences that require additional effort to resolve. In this work, for the first time, we synergize information from image, text, and event-data domains and introduce \textbf{OpenESS} to enable scalable ESS in an open-world, annotation-efficient manner. We achieve this goal by transferring the semantically rich CLIP knowledge from image-text pairs to event streams. To pursue better cross-modality adaptation, we propose a frame-to-event contrastive distillation and a text-to-event semantic consistency regularization. Experimental results on popular ESS benchmarks showed our approach outperforms existing methods. Notably, we achieve 53.93\% and 43.31\% mIoU on DDD17 and DSEC-Semantic without using either event or frame labels.
   \vspace{-1cm}
\end{abstract}

\section{Introduction}
\label{sec:intro}
Event cameras, often termed bio-inspired vision sensors, stand distinctively apart from traditional frame-based cameras and are often merited by their low latency, high dynamic range, and low power consumption \cite{huang2023event,son2017event,gehrig2022survey}. The realm of event-based vision perception, though nascent, has rapidly evolved into a focal point of contemporary research \cite{xu2023deep}. Drawing parallels with frame-based perception and recognition methodologies, a plethora of task-specific applications leveraging event cameras have burgeoned \cite{gallego2022event}.

Event-based semantic segmentation (ESS) emerges as one of the core event perception tasks and has gained increasing attention \cite{alonso2019ev-segnet,sun2022ess,hamaguchi2023hmnet,biswas2022halsie}. ESS inherits the challenges of traditional image segmentation \cite{long2015fully,cordts2016cityscapes,chen2017deeplab,chen2018encoder,he2016deep}, while also contending with the unique properties of event data \cite{alonso2019ev-segnet}, which opens up a plethora of opportunities for exploration. Although accurate and efficient dense predictions from event cameras are desirable for practical applications, the learning and annotation of the sparse, asynchronous, and high-temporal-resolution event streams pose several challenges \cite{kim2022spiking,messikommer2022ev-transfer,jia2023evsegformer}. Stemming from the image segmentation community, existing ESS models are trained on \textit{densely annotated} events within a \textit{fixed} and \textit{limited} set of label mapping \cite{alonso2019ev-segnet,sun2022ess}. Such closed-set learning from expensive annotations inevitably constrains the scalability of ESS systems.

An obvious approach will be to make use of the image domain and transfer knowledge to event data for the same vision tasks. Several recent attempts \cite{messikommer2022ev-transfer,sun2022ess,gehrig2020vid2e} resort to unsupervised domain adaptation to avoid the need for paired image and event data annotations for training. These methods demonstrate the potential of leveraging frame annotations to train a segmentation model for event data. However, transferring knowledge across frames and events is not straightforward and requires intermediate representations such as voxel grids, frame-like reconstructions, and bio-inspired spikes. Meanwhile, it is also costly to annotate dense frame labels for training, which limits their usage.

A recent trend inclines to the use of multimodal foundation models \cite{radford2021clip,kirillov2023segment,chen2023towards,peng2023learning,xu2024visual} to train task-specific models in an open-vocabulary and zero-shot manner, removing dependencies on human annotations. This paper continues such a trend. We propose a novel open-vocabulary framework for ESS, aiming at transferring pre-trained knowledge from both image and text domains to learn better representations of event data for the dense scene understanding task. Observing the large domain gap in between heterogeneous inputs, we design two cross-modality representation learning objectives that gradually align the event streams with images and texts. As shown in \cref{fig:teaser}, given raw events and text prompts as the input, the learned feature representations from our OpenESS framework exhibit promising results for known and unknown class segmentation and can be extended to more open-ended texts such as \textit{``adjectives"}, \textit{``fine-grained"}, and \textit{``coarse-grained"} descriptions.

To sum up, this work poses key contributions as follows:
\begin{itemize}
    \item We introduce OpenESS, a versatile event-based semantic segmentation framework capable of generating open-world dense event predictions given arbitrary text queries.
    \item To the best of our knowledge, this work represents the first attempt at distilling large vision-language models to assist event-based semantic scene understanding tasks.
    \item We propose a frame-to-event (F2E) contrastive distillation and a text-to-event (T2E) consistency regularization to encourage effective cross-modality knowledge transfer.
    \item Our approach sets up a new state of the art in annotation-free, annotation-efficient, and fully-supervised ESS settings on \textit{DDD17-Seg} and \textit{DSEC-Semantic} benchmarks.
\end{itemize}

\section{Related Work}
\label{sec:related_work}

\noindent\textbf{Event-based Vision.}
The microsecond-level temporal resolution, high dynamic range (typically $140$ dB \textit{vs.} $60$ dB of standard cameras), and power consumption efficiency of event cameras have posed a paradigm shift from traditional frame-based imaging \cite{gallego2022event,maqueda2018event,zou2022towards,steffen2019neuromorphic}. A large variety of event-based recognition, perception, localization, and reconstruction tasks have been established, encompassing object recognition \cite{gehrig2019representation,kim2021n-imagenet,cho2023label,peng2023get}, object detection \cite{gehrig2022dsec-det,zhou2023rgb,gehrig2023recurrent,zubić2023from}, depth estimation \cite{hidalgo2020learning,ghosh2020multi,mostafavi2021event,cho2023learning,pan2023srfnet,rançon2022stereospike}, optical flow \cite{cuadrado2023optical,zhu2019unsupervised,gehrig2021e-raft,brebion2021real,wan2022learning,gehrig2022dense,li2023blinkflow}, intensity-image reconstruction \cite{rebecq2019e2vid,ercan2023hyperE2VID,zhu2022event,ercan2023evreal,zhang2023formulating}, visual odometry and SLAM \cite{rebecq2016evo,hidalgo2022event,liu2021spatiotemporal}, stereoscopic panoramic imaging \cite{schraml2015stereo,belbachir2014hdr}, \etc. In this work, we focus on the recently-emerged task of event-based semantic scene understanding \cite{alonso2019ev-segnet,sun2022ess}. Such a pursuit is anticipated to tackle sparse, asynchronous, and high-temporal-resolution events for dense predictions, which is crucial for safety-critical in-drone or in-vehicle perceptions.

\noindent\textbf{Event-based Semantic Segmentation.}
The focus of ESS is on categorizing events into semantic classes for enhancing scene interpretation. Alonso \etal \cite{alonso2019ev-segnet} contributed the first benchmark based on DDD17 \cite{binas2017ddd17}. Subsequent works are tailored to improve the accuracy while mitigating the need for extensive event annotations \cite{gehrig2020vid2e}. EvDistill \cite{wang2021evdistill} and DTL \cite{wang2021dtl} utilized aligned frames to enhance event-based learning. EV-Transfer \cite{messikommer2022ev-transfer} and ESS \cite{sun2022ess} leveraged domain adaptation to transfer knowledge from existing image datasets to events. Recently, HALSIE \cite{biswas2022halsie} and HMNet \cite{hamaguchi2023hmnet} innovated ESS in cross-domain feature synthesis and memory-based event encoding. Another line of research pursues to use of spiking neural networks for energy-efficient ESS \cite{kim2022spiking,neftci2019surrogate,wu2019spatio,che2022differentiable}. In this work, different from previous pursuits, we aim to train ESS models in an annotation-free manner by distilling pre-trained vision-language models, hoping to address scalability and annotation challenges.

\begin{figure*}[t]
    \begin{center}
    \includegraphics[width=1.0\linewidth]{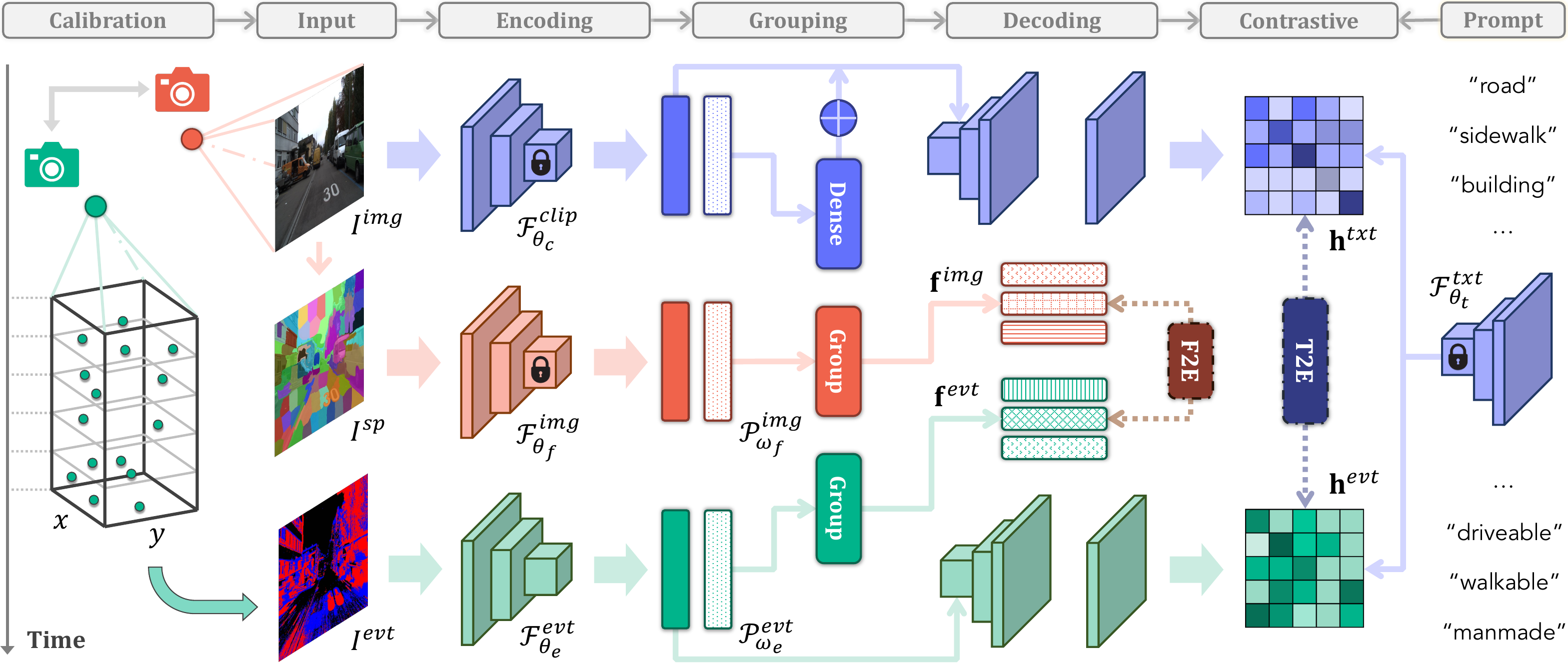}
    \end{center}
    \vspace{-0.4cm}
    \caption{\textbf{Architecture overview of the OpenESS framework}. We distill off-the-shelf knowledge from vision-languages models to event representations (\cf \cref{sec:revisit}). Given a calibrated event $I^{evt}$ and a frame $I^{img}$, we extract their features from the event network $\mathcal{F}^{evt}_{\theta_{e}}$ and the densified CLIP's image encoder $\mathcal{F}^{clip}_{\theta_{c}}$, which are then combined with the text embedding from CLIP's text encoder $\mathcal{F}^{txt}_{\theta_{t}}$ for open-world prediction (\cf \cref{sec:ov-ess}). To better serve for cross-modality knowledge transfer, we propose a \textbf{frame-to-event (F2E)} contrastive objective (\cf \cref{sec:f2e}) via superpixel-driven distillation and a \textbf{text-to-event (T2E)} consistency objective (\cf \cref{sec:t2e}) via scene-level regularization.}
    \label{fig:framework}
\end{figure*}

\noindent\textbf{Open-Vocabulary Learning.}
Recent advances in vision-language models open up new possibilities for visual perceptions \cite{wu2022survey,zhu2023survey,chen2023towards}. Such trends encompass image-based zero-shot and open-vocabulary detection \cite{li2022grounded,gao2022open,yao2023detclipv2,wu2023aligning}, as well as semantic \cite{zhou2022extract,li2022language,yu2023fc,ghiasi2022scaling,liang2023open}, instance \cite{huynh2022open,wu2023instance}, and panoptic \cite{he2023primitive,ding2022open,xu2023open} segmentation.
As far as we know, only three works studied the adaptation of CLIP for event-based recognition. EventCLIP \cite{wu2023eventclip} proposed to convert events to a 2D grid map and use an adapter to align event features with CLIP's knowledge. E-CLIP \cite{zhou2023e-clip} uses a hierarchical triple contrastive alignment that jointly unifies the event, image, and text feature embedding. Ev-LaFOR \cite{cho2023label} designed category-guided attraction and category-agnostic repulsion losses to bridge event with CLIP. Differently, we present the first attempt at adapting CLIP for dense predictions on sparse and asynchronous event streams. Our work is also close to superpixel-driven contrastive learning \cite{henaff2021efficient,sautier2022slidr}, where pre-processed superpixels are used to establish contrastive objectives with modalities from other tasks, \eg, point cloud understanding \cite{liu2023seal}, remote sensing \cite{guan2023pixel}, medical imaging \cite{wang2022separated}, and so on. In this work, we propose OpenESS to explore superpixel-to-event representation learning. Extensive experiments verify that such an approach is promising for annotation-efficient ESS.

\section{Methodology}
\label{sec:approach}

Our study serves as an early attempt at leveraging vision-language foundation models like CLIP \cite{radford2021clip} to learn meaningful event representations without accessing ground-truth labels. We start with a brief introduction of the CLIP model (\cf Sec.~\ref{sec:revisit}), followed by a detailed elaboration on our proposed open-vocabulary ESS (\cf \cref{sec:ov-ess}). To encourage effective cross-modal event representation learning, we introduce a frame-to-event contrastive distillation (\cf \cref{sec:f2e}) and a text-to-event consistency regularization (\cf \cref{sec:t2e}). An overview of the OpenESS framework is shown in \cref{fig:framework}.
 
\subsection{Revisiting CLIP}
\label{sec:revisit}

CLIP \cite{radford2021clip} learns to associate images with textual descriptions through a contrastive learning framework. It leverages a dataset of $400$ million image-text pairs, training an image encoder (based on a ResNet \cite{he2016deep} or Vision Transformer \cite{dosovitskiy2021vit}) and a text encoder (using a Transformer architecture \cite{vaswani2017attention}) to project images and texts into a shared embedding space. Such a training paradigm enables CLIP to perform zero-shot classification tasks, identifying images based on textual descriptions without specific training on those categories. To achieve annotation-free classification on a custom dataset, one needs to combine class label mappings with hand-crafted text prompts as the input to generate the text embedding. In this work, we aim to leverage the semantically rich CLIP feature space to assist open-vocabulary dense prediction on 
sparse and asynchronous event streams.

\subsection{Open-Vocabulary ESS}
\label{sec:ov-ess}

\noindent\textbf{Inputs.} Given a set of $N$ event data acquired by an event camera, we aim to segment each event $\mathbf{e}_i$ among the temporally ordered event streams $\varepsilon_{i}$, which are encoded by the pixel coordinates $(\mathbf{x}_i, \mathbf{y}_i)$, microsecond-level timestamp $t_{i}$, and the polarity $p_{i}\in\{-1, +1\}$ which indicates either an increase or decrease of the brightness. Each event camera pixel generates a spike whenever it perceives a change in logarithmic brightness that surpasses a predetermined threshold. Meanwhile, a conventional camera captures gray-scale or color frames $I^{img}_i\in\mathbb{R}^{3\times H\times W}$ which are spatially aligned and temporally synchronized with the events or can be aligned and synchronized to events via sensor calibration, where $H$ and $W$ are the spatial resolutions.

\noindent\textbf{Event Representations.}
Due to the sparsity, high temporal resolution, and asynchronous nature of event streams, it is common to convert raw events $\varepsilon_i$ into more regular representations $I^{evt}_i\in\mathbb{R}^{C\times H\times W}$ as the input to the neural network \cite{gallego2022event}, where $C$ denotes the number of embedding channels which is depended on the event representations themselves. Some popular choices of such embedding include spatiotemporal voxel grids \cite{zhu2018unsupervised,zhu2019unsupervised,gehrig2019representation}, frame-like reconstructions \cite{rebecq2019e2vid}, and bio-inspired spikes \cite{kim2022spiking}.  We investigate these three methods and show an example of taking voxel grids as the input in Fig.~\ref{fig:framework}. More analyses and comparisons using reconstructions and spikes are in later sections. Specifically, with a predefined number of events, each voxel grid is built from non-overlapping windows as:
\begin{equation}
   I^{evt}_i = \sum_{\mathbf{e}_j\in \varepsilon_i} p_j \delta(\mathbf{x}_j - \mathbf{x}) \delta(\mathbf{y}_j - \mathbf{y}) \max\{1 - |t^{*}_j - t| , 0\},~
\end{equation}
where $\delta$ is the Kronecker delta function; $t^{*}_j = (B-1)\frac{t_j - t_0}{\Delta T}$ is the normalized event timestamp with $B$ as the number of temporal bins in an event stream; $\Delta T$ is the time window and $t_0$ denotes the time of the first event in the window.

\noindent\textbf{Cross-Modality Encoding.}
Let $\mathcal{F}^{evt}_{\theta_{e}}:\mathbb{R}^{C\times H\times W}\mapsto\mathbb{R}^{D_1\times H_1\times W_1}$ be an event-based segmentation network with trainable parameters $\theta_e$, which takes as input an event embedding $I^{evt}_i$ and outputs a $D_1$-dimensional feature of downsampled spatial sizes $H_1$ and $W_1$. Meanwhile, we integrate CLIP's image encoder $\mathcal{F}^{clip}_{\theta_{c}}:\mathbb{R}^{3\times H\times W}\mapsto\mathbb{R}^{D_2\times H_2\times W_2}$ into our framework and keep the parameters $\theta_c$ fixed. The output is a $D_2$-dimensional feature of sizes $H_2$ and $W_2$. Our motivation is to transfer general knowledge from $\mathcal{F}^{clip}_{\theta_{c}}$ to $\mathcal{F}^{evt}_{\theta_{e}}$, such that the event branch can learn useful representations without using dense event annotations. To enable open-vocabulary ESS predictions, we leverage CLIP's text encoder $\mathcal{F}^{txt}_{\theta_{t}}$ with pre-trained parameters $\theta_{t}$. The input of $\mathcal{F}^{txt}_{\theta_{t}}$ comes from predefined text prompt templates and the output will be a text embedding extracted from CLIP's rich semantic space.

\noindent\textbf{Densifications.}
CLIP was originally designed for image-based recognition tasks and does not provide per-pixel outputs for dense predictions. Several recent attempts explored the adaptation from global, image-level recognition to local, pixel-level prediction, via either model structure modification \cite{zhou2022extract} or fine-tuning \cite{rao2022dense-clip,yu2023fc,li2022language}. The former directly reformulates the value-embedding layer in CLIP's image encoder, while the latter uses semantic labels to gradually adapt the pre-trained weights to generate dense predictions. In this work, we implement both solutions to densify CLIP's outputs and compare their performances in our experiments.

Up until now, we have presented a preliminary framework capable of conducting open-vocabulary ESS by leveraging knowledge from the CLIP model. However, due to the large domain gap between the event and image modalities, a na\"{\i}ve adaptation is sub-par in tackling the challenging event-based semantic scene understanding task.

\subsection{F2E: Frame-to-Event Contrastive Distillation}
\label{sec:f2e}

Since our objective is to encourage effective cross-modality knowledge transfer for holistic event scene perception, it thus becomes crucial to learn meaningful representations for both \textit{thing} and \textit{stuff} classes, especially their boundary information. However, the sparsity and asynchronous nature of event streams inevitably impede such objectives.

\noindent\textbf{Superpixel-Driven Knowledge Distillation.}
To pursue a more informative event representation learning at higher granularity, we propose to first leverage calibrated frames to generate coarse, instance-level superpixels and then distill knowledge from a pre-trained image backbone to the event segmentation network. Superpixel groups pixels into conceptually meaningful atomic regions, which can be used as the basis for higher-level perceptions \cite{wang2011superpixel-tracking,achanta2012slic,li2015superpixel}. The semantically coherent frame-to-event correspondences can thus be found using pre-processed or online-generated superpixels. Such correspondences tend to bridge the sparse events to dense frame pixels in a holistic manner without involving extra training or annotation efforts.

\noindent\textbf{Superpixel \& Superevent Generation.}
We resort to the following two ways of generating the superpixels. The first way is to leverage heuristic methods, \eg SLIC \cite{achanta2012slic}, to efficiently groups pixels from frame $I^{img}_i$ into a total of $M_{slic}$ segments with good boundary adherence and regularity as $I^{sp}_i=\{\mathcal{I}_i^1, \mathcal{I}_i^2, ..., \mathcal{I}_i^{M_{slic}}\}$, where $M_{slic}$ is a hyperparameter that needs to be adjusted based on the inputs. The generated superpixels satisfy $\mathcal{I}_i^1\cup \mathcal{I}_i^2 \cup ... \cup \mathcal{I}_i^{M_{slic}} = \{ 1, 2, ..., H\times W\}$. 
For the second option, we use the recent Segment Anything Model (SAM) \cite{kirillov2023segment} which takes $I^{img}_i$ as the input and outputs $M_{sam}$ class-agnostic masks. For simplicity, we use $M$ to denote the number of superpixels used during knowledge distillation, \ie, $\{I^{sp}_i=\{\mathcal{I}_i^1,...,\mathcal{I}_i^k\}|k=1,...,M\}$ and show more comparisons between SLIC \cite{achanta2012slic} and SAM \cite{kirillov2023segment} in later sections.
Since $I^{evt}_i$ and $I^{img}_i$ have been aligned and synchronized, we can group events from $I^{evt}_i$ into superevents $\{V^{sp}_i=\{\mathcal{V}_i^1,...,\mathcal{V}_i^l\}|l=1,...,M\}$ by using the known event-pixel correspondences. 

\noindent\textbf{Frame-to-Event Contrastive Learning.}
To encourage better superpixel-level knowledge transfer, we leverage a pre-trained image network $\mathcal{F}^{img}_{\theta_{f}}:\mathbb{R}^{3\times H\times W}\mapsto\mathbb{R}^{D_3\times H_3\times W_3}$ as the teacher and distill information from it to the event branch $\mathcal{F}^{evt}_{\theta_{e}}$. The parameters of $\mathcal{F}^{img}_{\theta_{f}}$, which can come from either CLIP \cite{radford2021clip} or other pretext task pre-trained backbones such as \cite{chen2020mocov2,oquab2023dinov2,caron2020swav}, are kept frozen during the distillation. With $\mathcal{F}^{evt}_{\theta_{e}}$ and $\mathcal{F}^{img}_{\theta_{f}}$, we generate the superevent and superpixel features as follows:
\begin{align}
    \mathbf{f}^{evt}_i =& ~\frac{1}{|V^{sp}_i|}\sum_{l\in V^{sp}_i}\mathcal{P}_{\omega_e}^{evt}~(~\mathcal{F}^{evt}_{\theta_{e}} ~(I^{evt}_i)_l~ )~,\\
    \mathbf{f}^{img}_i =& ~\frac{1}{|I^{sp}_i|}\sum_{k\in I^{sp}_i}\mathcal{P}_{\omega_f}^{img}~(~\mathcal{F}^{img}_{\theta_{f}} ~(I^{img}_i)_k~ )~,
\end{align}
where $\mathcal{P}_{\omega_e}^{evt}$ and $\mathcal{P}_{\omega_f}^{img}$ are projection layers with trainable parameters $\omega_e$ and $\omega_f$, respectively, for the event branch and frame branch. In the actual implementation, $\mathcal{P}_{\omega_e}^{evt}$ and $\mathcal{P}_{\omega_f}^{img}$ consist of linear layers which map the $D_1$- and $D_3$-dimensional event and frame features to the same shape. The following contrastive learning objective is applied to the event prediction and the frame prediction:
\begin{align}
    \mathcal{L}_{F2E}(\theta_{e},\omega_e,\omega_f) = -\sum_i \log \left[ \frac{e^{(\langle \mathbf{f}^{evt}_i, \mathbf{f}^{img}_i \rangle/\tau_1 )}}{\sum_{j\neq i} e^{(\langle \mathbf{f}^{evt}_i, \mathbf{f}^{img}_j \rangle/\tau_1 )}} \right]~,
    \label{eq:f2e}
\end{align}
where $\langle \cdot,\cdot\rangle$ denotes the scalar product between the superevent and superpixel embedding; $\tau_1>0$ is a temperature coefficient that controls the pace of knowledge transfer.

\noindent\textbf{Role in Our Framework.}
Our F2E contrastive distillation establishes an effective pipeline for transferring superpixel-level knowledge from dense, visual informative frame pixels to sparse, irregular event streams. Since we are targeting the semantic segmentation task, the learned event representations should be able to reason in terms of instances and instance parts at and in between semantic boundaries.

\subsection{T2E: Text-to-Event Consistency Regularization}
\label{sec:t2e}
Although the aforementioned frame-to-event knowledge transfer provides a simple yet effective way of transferring off-the-shelf knowledge from frames to events, the optimization objective might encounter unwanted conflicts.

\noindent\textbf{Intra-Class Optimization Conflict.} During the model pre-training, the superpixel-driven contrastive loss takes the corresponding superevent and superpixel pair in a batch as the positive pair, while treating all remaining pairs as negative samples. Since heuristic superpixels only provide a coarse grouping of conceptually coherent segments (kindly refer to our Appendix for more detailed analysis), it is thus inevitable to encounter self-conflict during the optimization. That is to say, from hindsight, there is a chance that the superpixels belonging to the same semantic class could be involved in both positive and negative samples.

\noindent\textbf{Text-Guided Semantic Regularization.}
To mitigate the possible self-conflict in \cref{eq:f2e}, we propose a text-to-event semantic consistency regularization mechanism that leverages CLIP's text encoder to generate semantically more consistent text-frame pairs $\{I_{i}^{img}, T_{i}\}$, where $T_{i}$ denotes the text embedding extracted from $\mathcal{F}^{txt}_{\theta_{t}}$. Such a paired relationship can be leveraged via CLIP without additional training. We then construct event-text pairs $\{I_{i}^{evt}, T_{i}\}$ by propagating the alignment between events and frames. Specifically, the paired event and text features are extracted as follows:
\begin{align}
    \mathbf{h}^{evt}_i =& ~\mathcal{Q}_{\omega_q}^{evt}~(\mathcal{F}^{evt}_{\theta_{e}} ~(I^{evt}_i) )~, ~~
    \mathbf{h}^{txt}_i = \mathcal{F}^{txt}_{\theta_{t}}~(T_{i} )~,
\end{align}
where $\mathcal{Q}_{\omega_q}^{evt}$ is a projection layer with trainable parameters $\omega_q$, which is similar to that of $\mathcal{P}_{\omega_e}^{evt}$. Now assume there are a total of $Z$ classes in the event dataset, the following objective is applied to encourage the consistency regularization:

\setlength{\abovedisplayskip}{-12pt}
\setlength{\abovedisplayshortskip}{0pt}
\begin{align}
    &\mathcal{L}_{T2E}(\theta_{e},\omega_q) =\\ &- \sum_{z=1}^Z \log \left[ \frac{\sum_{T_i \in z,I^{evt}_i} e^{(\langle \mathbf{h}^{evt}_i, \mathbf{h}^{txt}_i \rangle/\tau_2 )}}{\sum_{j\neq i, T_i \in z,T_i \not\in I^{evt}_i} e^{(\langle \mathbf{h}^{evt}_j, \mathbf{h}^{txt}_i \rangle/\tau_2 )}} \right]~,
    \label{eq:t2e}
\end{align}
where $\tau_2>0$ is a temperature coefficient that controls the pace of knowledge transfer.
The overall optimization objective of our OpenESS framework is to minimize $\mathcal{L}=\mathcal{L}_{F2E} + \alpha \mathcal{L}_{T2E}$, where $\alpha$ is a weight balancing coefficient.

\noindent\textbf{Role in Our Framework.}
Our T2E semantic consistency regularization provides a global-level alignment to compensate for the possible self-conflict in the superpixel-driven frame-to-event contrastive learning. As we will show in the following sections, the two objectives work synergistically in improving the performance of open-vocabulary ESS.

\noindent\textbf{Inference-Time Configuration.}
Our OpenESS framework is designed to pursue segmentation accuracy in annotation-free and annotation-efficient manners, without sacrificing event processing efficiency. As can be seen from \cref{fig:framework}, after the cross-modality knowledge transfer, only the event branch will be kept. This guarantees that there will be no extra latency or power consumption added during the inference, which is in line with the practical requirements.

\section{Experiments}
\label{sec:experiments}

\begin{table}[t]
\caption{\textbf{Comparative study} of existing ESS approaches under the annotation-free, fully-supervised, and open-vocabulary ESS settings, respectively, on the \textit{test} sets of the \textit{DDD17-Seg} \cite{binas2017ddd17} and \textit{DSEC-Semantic} \cite{sun2022ess} datasets. All scores are in percentage ($\%$). The \textbf{best} score from each learning setting is highlighted in \textbf{bold}.}
\vspace{-0.1cm}
\centering\scalebox{0.79}{
\begin{tabular}{r|r|p{27.5pt}<{\centering}p{27.5pt}<{\centering}|p{27.5pt}<{\centering}p{27.5pt}<{\centering}}
    \toprule
    \multirow{2}{*}{\textbf{Method}} & \multirow{2}{*}{\textbf{Venue}} & \multicolumn{2}{c|}{\textbf{DDD17}} & \multicolumn{2}{c}{\textbf{DSEC}}
    \\
    & & Acc & mIoU & Acc & mIoU
    \\\midrule\midrule
    \rowcolor{robo_blue!7}\multicolumn{6}{l}{\textcolor{robo_blue}{\textbf{Annotation-Free ESS}}}
    \\
    MaskCLIP \cite{zhou2022extract} & {\small ECCV'22} & $81.29$ & $31.90$ & $58.96$ & $21.97$
    \\
    FC-CLIP \cite{yu2023fc} & {\small NeurIPS'23} & $88.66$ & $51.12$ & $79.20$ & $39.42$
    \\
    \textbf{OpenESS} & \textbf{Ours} & \cellcolor{robo_blue!7}{$\mathbf{90.51}$} & \cellcolor{robo_blue!7}{$\mathbf{53.93}$} & \cellcolor{robo_blue!7}{$\mathbf{86.18}$} & \cellcolor{robo_blue!7}{$\mathbf{43.31}$}
    \\\midrule
    \rowcolor{robo_red!7}\multicolumn{6}{l}{\textcolor{robo_red}{\textbf{Fully-Supervised ESS}}}
    \\
    Ev-SegNet \cite{alonso2019ev-segnet} & {\small CVPRW'19} & $89.76$ & $54.81$ & $88.61$ & $51.76$
    \\
    E2VID \cite{rebecq2019e2vid} & {\small TPAMI'19} & $85.84$ & $48.47$ & $80.06$ & $44.08$
    \\
    Vid2E \cite{gehrig2020vid2e} & {\small CVPR'20} & $90.19$ & $56.01$ & - & -
    \\
    EVDistill \cite{wang2021evdistill} & {\small CVPR'21} & - & $58.02$ & - & -
    \\
    DTL \cite{wang2021dtl} & {\small ICCV'21} & - & $58.80$ & - & -
    \\
    PVT-FPN \cite{wang2021pvt} & {\small ICCV'21} & $94.28$ & $53.89$ & - & -
    \\
    SpikingFCN \cite{kim2022spiking} & {\small NCE'22} & - & $34.20$ & - & -
    \\
    EV-Transfer \cite{messikommer2022ev-transfer} & {\small RA-L'22} & $51.90$ & $15.52$ & $63.00$ & $24.37$
    \\
    ESS \cite{sun2022ess} & {\small ECCV'22} & $88.43$ & $53.09$ & $84.17$ & $45.38$
    \\
    ESS-Sup \cite{sun2022ess} & {\small ECCV'22} & $91.08$ & \cellcolor{robo_red!7}{$\mathbf{61.37}$} & $89.37$ & $53.29$
    \\
    P2T-FPN \cite{wu2023p2t} & {\small TPAMI'23} & $94.57$ & $54.64$ & - & -
    \\
    EvSegformer \cite{jia2023evsegformer} & {\small TIP'23} & \cellcolor{robo_red!7}{$\mathbf{94.72}$} & $54.41$ & - & - 
    \\
    HMNet-B \cite{hamaguchi2023hmnet} & {\small CVPR'23} & - & - & $88.70$ & $51.20$ 
    \\
    HMNet-L \cite{hamaguchi2023hmnet} & {\small CVPR'23} & - & - & \cellcolor{robo_red!7}{$\mathbf{89.80}$} & \cellcolor{robo_red!7}{$\mathbf{55.00}$}
    \\
    HALSIE \cite{biswas2022halsie} & {\small WACV'24} & $92.50$ & $60.66$ & $89.01$ & $52.43$
    \\\midrule
    \rowcolor{robo_green!7}\multicolumn{6}{l}{\textcolor{robo_green}{\textbf{Open-Vocabulary ESS}}}
    \\
    MaskCLIP \cite{zhou2022extract} & {\small ECCV'22} & $90.50$ & $61.27$ & $89.81$ & $55.01$
    \\
    FC-CLIP \cite{yu2023fc} & {\small NeurIPS'23} & $90.68$ & $62.01$ & $89.97$ & $55.67$
    \\
    \textbf{OpenESS} & \textbf{Ours} & \cellcolor{robo_green!7}{$\mathbf{91.05}$} & \cellcolor{robo_green!7}{$\mathbf{63.00}$} & \cellcolor{robo_green!7}{$\mathbf{90.21}$} & \cellcolor{robo_green!7}{$\mathbf{57.21}$}
    \\\bottomrule
\end{tabular}}
\label{tab:comparative}
\end{table}

\subsection{Settings}
\noindent\textbf{Datasets.} We conduct experiments on two popular ESS datasets. \textit{\textbf{DDD17-Seg}} \cite{alonso2019ev-segnet} is a widely used ESS benchmark consisting of $40$ sequences acquired by a DAVIS346B. In total, $15950$ training and $3890$ testing events of spatial size $352\times 200$ are used, along with synchronized gray-scale frames provided by the DAVIS camera.
\textit{\textbf{DSEC-Semantic}} \cite{sun2022ess} provides semantic labels for $11$ sequences in the DSEC \cite{gehrig2021dsec} dataset. The training and testing splits contain $8082$ and $2809$ events of spatial size $640\times 440$, accompanied by color frames (with sensor calibration parameters available) recorded at $20$Hz. More details are in the Appendix.

\noindent\textbf{Benchmark Setup.} In addition to the conventional fully-supervised ESS, we establish two open-vocabulary ESS settings for \textit{annotation-free} and \textit{annotation-efficient} learning, respectively. The former aims to train an ESS model without using any dense event labels, while the latter assumes an annotation budget of 1\%, 5\%, 10\%, or 20\% of events in the training set. We treat the first few samples from each sequence as labeled and the remaining ones as unlabeled.

\noindent\textbf{Implementation Details.} Our framework is implemented using PyTorch \cite{paszke2019pytorch}. Based on the use of event representations, we form \texttt{frame2voxel}, \texttt{frame2recon}, and \texttt{frame2spike} settings, where the event branch will adopt E2VID \cite{rebecq2019e2vid}, ResNet-50 \cite{he2016deep}, and SpikingFCN \cite{kim2022spiking}, respectively, with an AdamW \cite{loshchilov2019adamw} optimizer with cosine learning rate scheduler. The frame branch uses a pre-trained ResNet-50 \cite{caron2021dino,chen2020mocov2,caron2020swav} and is kept frozen. The number of superpixels involved in the calculation of F2E contrastive loss is set to $100$ for \textit{DSEC-Semantic} \cite{sun2022ess} and $25$ for \textit{DDD17-Seg} \cite{alonso2019ev-segnet}. For evaluation, we extract the feature embedding for each text prompt offline from a frozen CLIP text encoder using pre-defined templates. For linear probing, the pre-trained event network $\mathcal{F}^{evt}_{\theta_{e}}$ is kept frozen, followed by a trainable point-wise linear classification head. Due to space limits, kindly refer to our Appendix for additional details.

\subsection{Comparative Study}

\noindent\textbf{Annotation-Free ESS.}
In \cref{tab:comparative}, we compare OpenESS with MaskCLIP \cite{zhou2022extract} and FC-CLIP \cite{yu2023fc} in the absence of event labels. Our approach achieves zero-shot ESS results of $53.93\%$ and $43.31\%$ on \textit{DDD17-Seg} \cite{alonso2019ev-segnet} and \textit{DSEC-Semantic} \cite{sun2022ess}, much higher than the two competitors and even comparable to some fully-supervised methods. This validates the effectiveness of conducting ESS in an annotation-free manner for practical usage. Meanwhile, we observe that a fine-tuned CLIP encoder \cite{yu2023fc} could generate much better semantic predictions than the structure adaptation method \cite{zhou2022extract}, as mentioned in \cref{sec:ov-ess}.

\noindent\textbf{Comparisons to State-of-the-Art Methods.}
As shown in \cref{tab:comparative}, the proposed OpenESS sets up several new state-of-the-art results in the two ESS benchmarks. Compared to the previously best-performing methods, OpenESS is $1.63\%$ and $2.21\%$ better in terms of mIoU scores on \textit{DDD17-Seg} \cite{alonso2019ev-segnet} and \textit{DSEC-Semantic} \cite{sun2022ess}, respectively. It is worth mentioning that in addition to the performance improvements, our approach can generate open-vocabulary predictions that are beyond the closed sets of predictions of existing methods, which is more in line with the practical usage.

\begin{figure}[t]
    \begin{center}
    \includegraphics[width=1.0\linewidth]{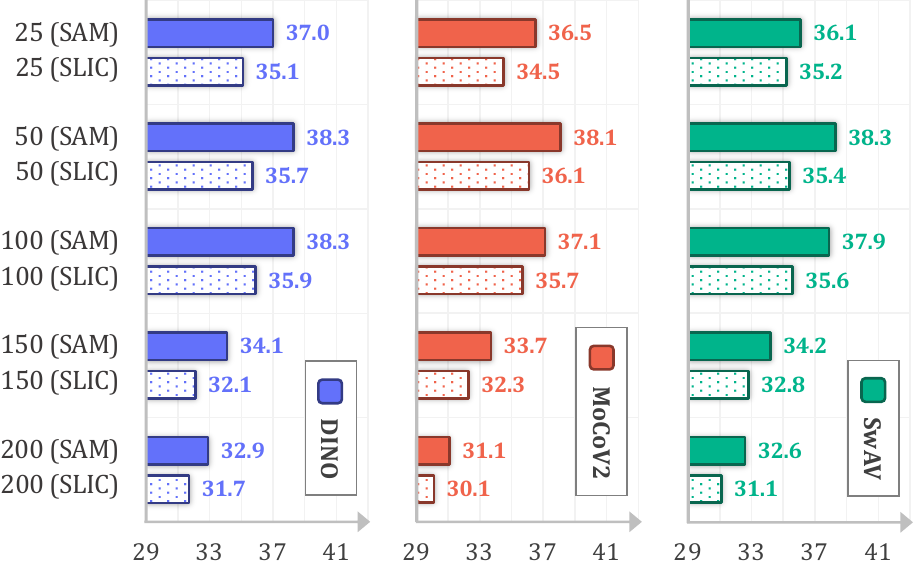}
    \end{center}
    \vspace{-0.42cm}
    \caption{\textbf{Ablation study} on the number of superpixels (provided by either SAM \cite{kirillov2023segment} or SLIC \cite{achanta2012slic}) involved in calculating the frame-to-event contrastive loss. Models after pre-training are fine-tuned with 1\% annotations. All mIoU scores are in percentage ($\%$).}
    \label{fig:ablation}
\end{figure}

\noindent\textbf{Annotation-Efficient Learning.}
We establish a comprehensive benchmark for ESS under limited annotation scenarios and show the results in \cref{tab:comparative_finetune}. As can be seen, the proposed OpenESS contributes significant performance improvements over random initialization under linear probing, few-shot fine-tuning, and fully-supervised learning settings. Specifically, using either voxel grid or event reconstruction representation, our approach achieves $>30\%$ relative gains in mIoU on both datasets under liner probing and around $2\%$ higher than prior art in mIoU with full supervisions. We also observe that using voxel grids to represent raw event streams tends to yield overall better ESS performance.

\begin{table}[t]
\caption{\textbf{Comparative study} of different representation learning methods applied on event data. \textbf{OV} denotes whether supporting open-vocabulary predictions. All mIoU scores are in percentage ($\%$). The \textbf{best} score from each dataset is highlighted in \textbf{bold}.}
\vspace{-0.1cm}
\centering\scalebox{0.79}{
\begin{tabular}{r|r|c|p{20pt}<{\centering}|p{34pt}<{\centering}|p{34pt}<{\centering}}
    \toprule
    \textbf{Method} & \textbf{Venue} & \textbf{Backbone} & \textbf{OV} & \textbf{DDD17} & \textbf{DSEC}
    \\\midrule\midrule
    \rowcolor{gray!9}\textcolor{gray}{Random} & \textcolor{gray}{-} & \textcolor{gray}{ViT-S/16} & \textcolor{gray}{\textbf{\ding{55}}} & \textcolor{gray}{$48.76$} & \textcolor{gray}{$40.53$}
    \\
    MoCoV3 \cite{chen2021mocov3} & {\small ICCV'21} & ViT-S/16 & \textcolor{robo_red}{\textbf{\ding{55}}} & $53.65$ & $49.21$
    \\
    IBoT \cite{zhou2021ibot} & {\small ICLR'22} & ViT-S/16 & \textcolor{robo_red}{\textbf{\ding{55}}} & $49.94$ & $42.53$
    \\
    ECDP \cite{yang2023ecdp} & {\small ICCV'23} & ViT-S/16 & \textcolor{robo_red}{\textbf{\ding{55}}} & $54.66$ & $47.91$
    \\
    \rowcolor{gray!9}\textcolor{gray}{Random} & \textcolor{gray}{-} & \textcolor{gray}{ViT-B/16} & \textcolor{gray}{\textbf{\ding{55}}} & \textcolor{gray}{$43.89$} & \textcolor{gray}{$38.24$}
    \\
    BeiT \cite{bao2021beit} & {\small ICLR'22} & ViT-B/16 & \textcolor{robo_red}{\textbf{\ding{55}}} & $52.39$ & $46.52$
    \\
    MAE \cite{he2022mae} & {\small CVPR'22} & ViT-B/16 & \textcolor{robo_red}{\textbf{\ding{55}}} & $52.36$ & $47.56$
    \\\midrule
    \rowcolor{gray!9}\textcolor{gray}{Random} & \textcolor{gray}{-} & \textcolor{gray}{ResNet-50} & \textcolor{gray}{\textbf{\ding{55}}} & \textcolor{gray}{$56.96$} & \textcolor{gray}{$57.60$}
    \\
    SimCLR \cite{chen2020simclr} & {\small ICML'20} & ResNet-50 & \textcolor{robo_red}{\textbf{\ding{55}}} & $57.22$ & $59.06$
    \\
    ECDP \cite{yang2023ecdp} & {\small ICCV'23} & ResNet-50 & \textcolor{robo_red}{\textbf{\ding{55}}} & $59.15$ & $\mathbf{59.16}$
    \\\midrule
    \rowcolor{gray!9}\textcolor{gray}{Random} & \textcolor{gray}{-} & \textcolor{gray}{ResNet-50} & \textcolor{gray}{\textbf{\ding{55}}} & \textcolor{gray}{$55.56$} & \textcolor{gray}{$52.86$}
    \\
    \textbf{OpenESS} & \textbf{Ours} & ResNet-50 & \textcolor{robo_green}{\textbf{\checkmark}} & $57.01$ & $55.01$
    \\
    \rowcolor{gray!9}\textcolor{gray}{Random} & \textcolor{gray}{-} & \textcolor{gray}{E2VID} & \textcolor{gray}{\textbf{\ding{55}}} & \textcolor{gray}{$61.06$} & \textcolor{gray}{$54.96$}
    \\
    \textbf{OpenESS} & \textbf{Ours} & E2VID & \textcolor{robo_green}{\textbf{\checkmark}} & $\mathbf{63.00}$ & $57.21$
    \\\bottomrule
\end{tabular}}
\label{tab:pretrain}
\end{table}
\begin{table*}[t]
\caption{\textbf{Comparative study} of different open-vocabulary semantic segmentation methods \cite{zhou2022extract,yu2023fc} under the linear probing (LP) and few-shot fine-tuning, and full supervision (Full) settings, respectively, on the \textit{test} sets of the \textit{DDD17-Seg} \cite{binas2017ddd17} and \textit{DSEC-Semantic} \cite{sun2022ess} datasets. All mIoU scores are given in percentage ($\%$). The \textbf{best} mIoU scores from each learning configuration are highlighted in \textbf{bold}.}
\vspace{-0.1cm}
\centering\scalebox{0.79}{
\begin{tabular}{r|c|cccccc|cccccc}
    \toprule
    \multirow{2}{*}{\textcolor{darkgray}{\textbf{Method}}} & \multirow{2}{*}{\textcolor{darkgray}{\textbf{Configuration}}} & \multicolumn{6}{c|}{\textcolor{darkgray}{\textbf{DSEC-Semantic}}} & \multicolumn{6}{c}{\textcolor{darkgray}{\textbf{DDD17-Seg}}}
    \\
    & & LP & 1\% & 5\% & 10\% & 20\% & Full & LP & 1\% & 5\% & 10\% & 20\% & Full
    \\\midrule\midrule
    \rowcolor{gray!9}\textcolor{gray}{Random} & \textcolor{gray}{Voxel Grid} & \textcolor{gray}{$6.70$} & \textcolor{gray}{$26.62$} & \textcolor{gray}{$31.22$} & \textcolor{gray}{$33.67$} & \textcolor{gray}{$41.31$} & \textcolor{gray}{$54.96$} & \textcolor{gray}{$12.30$} & \textcolor{gray}{$52.13$} & \textcolor{gray}{$54.87$} & \textcolor{gray}{$58.66$} & \textcolor{gray}{$59.52$} & \textcolor{gray}{$61.06$}
    \\\midrule
    MaskCLIP \cite{zhou2022extract} &  & $33.08$ & $33.89$ & $37.03$ & $38.83$ & $42.40$ & $55.01$ & $31.91$ & $53.91$ & $56.27$ & $59.32$ & $59.97$ & $61.27$
    \\
    FC-CLIP \cite{yu2023fc} & Voxel Grid & $43.00$ & $39.12$ & $43.71$ & $44.09$ & $47.77$ & $55.67$ & $54.07$ & $56.38$ & $58.50$ & $60.05$ & $60.85$ & $62.01$
    \\
    \textbf{OpenESS (Ours)} & \textcolor{robo_blue}{\textbf{\texttt{frame2voxel}}} & $\mathbf{44.26}$ & $\mathbf{41.41}$ & $\mathbf{44.97}$ & $\mathbf{46.25}$ & $\mathbf{48.28}$ & $\mathbf{57.21}$ & $\mathbf{55.61}$ & $\mathbf{57.58}$ & $\mathbf{59.07}$ & $\mathbf{61.03}$ & $\mathbf{61.78}$ & $\mathbf{63.00}$
    \\
    \textit{Improve}~$\uparrow$ & & {\small\textcolor{robo_blue}{$+33.56$}} & {\small\textcolor{robo_blue}{$+14.79$}} & {\small\textcolor{robo_blue}{$+13.75$}} & {\small\textcolor{robo_blue}{$+12.58$}} & {\small\textcolor{robo_blue}{$+6.97$}} & {\small\textcolor{robo_blue}{$+2.25$}} & {\small\textcolor{robo_blue}{$+43.31$}} & {\small\textcolor{robo_blue}{$+5.45$}} & {\small\textcolor{robo_blue}{$+4.20$}} & {\small\textcolor{robo_blue}{$+2.37$}} & {\small\textcolor{robo_blue}{$+2.26$}} & {\small\textcolor{robo_blue}{$+1.94$}}
    \\\midrule\midrule
    \rowcolor{gray!9}\textcolor{gray}{Random} & \textcolor{gray}{Reconstruction} & \textcolor{gray}{$6.22$} & \textcolor{gray}{$23.95$} & \textcolor{gray}{$30.42$} & \textcolor{gray}{$34.11$} & \textcolor{gray}{$39.25$} & \textcolor{gray}{$52.86$} & \textcolor{gray}{$13.89$} & \textcolor{gray}{$45.30$} & \textcolor{gray}{$52.03$} & \textcolor{gray}{$53.02$} & \textcolor{gray}{$54.05$} & \textcolor{gray}{$55.56$}
    \\\midrule
    MaskCLIP \cite{zhou2022extract} & & $27.09$ & $30.73$ & $36.33$ & $40.13$ & $43.37$ & $52.97$ & $29.81$ & $49.02$ & $53.65$ & $54.11$ & $54.75$ & $56.12$
    \\
    FC-CLIP \cite{yu2023fc} & Reconstruction & $40.08$ & $38.99$ & $43.34$ & $45.35$ & $47.18$ & $53.05$ & $52.17$ & $51.01$ & $54.09$ & $54.99$ & $55.05$ & $56.34$
    \\
    \textbf{OpenESS (Ours)} & \textcolor{robo_red}{\textbf{\texttt{frame2recon}}} & $\mathbf{44.08}$ & $\mathbf{43.17}$ & $\mathbf{45.58}$ & $\mathbf{48.94}$ & $\mathbf{49.74}$ & $\mathbf{55.01}$ & $\mathbf{53.61}$ & $\mathbf{52.02}$ & $\mathbf{55.11}$ & $\mathbf{55.66}$ & $\mathbf{56.07}$ & $\mathbf{57.01}$
    \\
    \textit{Improve}~$\uparrow$ & & {\small\textcolor{robo_red}{$+37.86$}} & {\small\textcolor{robo_red}{$+19.22$}} & {\small\textcolor{robo_red}{$+15.16$}} & {\small\textcolor{robo_red}{$+14.83$}} & {\small\textcolor{robo_red}{$+10.49$}} & {\small\textcolor{robo_red}{$+2.15$}} & {\small\textcolor{robo_red}{$+39.72$}} & {\small\textcolor{robo_red}{$+6.72$}} & {\small\textcolor{robo_red}{$+3.08$}} & {\small\textcolor{robo_red}{$+2.64$}} & {\small\textcolor{robo_red}{$+2.02$}} & {\small\textcolor{robo_red}{$+1.45$}}
    \\\bottomrule
\end{tabular}}
\label{tab:comparative_finetune}
\vspace{-0.2cm}
\end{table*}
\begin{table}[t]
\caption{\textbf{Ablation study} of OpenESS under linear probing (LP) and few-shot fine-tuning settings from three learning configurations on the \textit{test} set of \textit{DDD17-Seg} \cite{binas2017ddd17}. \textbf{F2E} denotes the frame-to-event contrastive learning. \textbf{T2E} denotes the text-to-event semantic regularization. All mIoU scores are given in percentage ($\%$).}
\vspace{-0.1cm}
\centering\scalebox{0.79}{
\begin{tabular}{c|cc|p{19.5pt}<{\centering}p{19.5pt}<{\centering}p{19.5pt}<{\centering}p{19.5pt}<{\centering}p{19.5pt}<{\centering}}
    \toprule
    \multirow{2}{*}{\textcolor{darkgray}{\textbf{Configuration}}} & \multirow{2}{*}{\textcolor{darkgray}{\textbf{F2E}}} & \multirow{2}{*}{\textcolor{darkgray}{\textbf{T2E}}} & \multicolumn{5}{c}{\textcolor{darkgray}{\textbf{DDD17-Seg}}}
    \\
    & & & LP & 1\% & 5\% & 10\% & 20\%
    \\\midrule\midrule
    \rowcolor{gray!9}Voxel Grid & \multicolumn{2}{c|}{\textcolor{gray}{Random}} & \textcolor{gray}{$12.30$} & \textcolor{gray}{$52.13$} & \textcolor{gray}{$54.87$} & \textcolor{gray}{$58.66$} & \textcolor{gray}{$59.52$}
    \\\midrule
    \multirow{3}{*}{\textcolor{robo_blue}{\textbf{\texttt{frame2voxel}}}} & \textcolor{robo_blue}{\textbf{\checkmark}} & & $52.60$ & $55.41$ & $57.07$ & $59.77$ & $60.21$
    \\
    & & \textcolor{robo_blue}{\textbf{\checkmark}} & $54.11$ & $56.77$ & $58.95$ & $60.12$ & $60.99$
    \\
    & \textcolor{robo_blue}{\textbf{\checkmark}} & \textcolor{robo_blue}{\textbf{\checkmark}} & $55.61$ & $57.58$ & $59.07$ & $61.03$ & $61.78$
    \\\midrule\midrule
    \rowcolor{gray!9}Reconstruction & \multicolumn{2}{c|}{\textcolor{gray}{Random}} & \textcolor{gray}{$13.89$} & \textcolor{gray}{$45.30$} & \textcolor{gray}{$52.03$} & \textcolor{gray}{$53.02$} & \textcolor{gray}{$54.05$}
    \\\midrule
    \multirow{3}{*}{\textcolor{robo_red}{\textbf{\texttt{frame2recon}}}} & \textcolor{robo_red}{\textbf{\checkmark}} & & $50.21$ & $50.96$ & $53.67$ & $54.21$ & $54.92$
    \\
    & & \textcolor{robo_red}{\textbf{\checkmark}} & $52.62$ & $51.63$ & $54.27$ & $55.00$ & $55.17$
    \\
    & \textcolor{robo_red}{\textbf{\checkmark}} & \textcolor{robo_red}{\textbf{\checkmark}} & $53.61$ & $52.02$ & $55.11$ & $55.66$ & $56.07$
    \\\midrule\midrule
    \rowcolor{gray!9}Spike & \multicolumn{2}{c|}{\textcolor{gray}{Random}} & \textcolor{gray}{$12.04$} & \textcolor{gray}{$10.01$} & \textcolor{gray}{$20.02$} & \textcolor{gray}{$25.81$} & \textcolor{gray}{$26.03$}
    \\\midrule
    \multirow{3}{*}{\textcolor{robo_green}{\textbf{\texttt{frame2spike}}}} & \textcolor{robo_green}{\textbf{\checkmark}} & & $15.07$ & $14.31$ & $21.77$ & $26.89$ & $27.07$
    \\
    & & \textcolor{robo_green}{\textbf{\checkmark}} & $16.11$ & $14.67$ & $22.61$ & $27.97$ & $29.01$
    \\
    & \textcolor{robo_green}{\textbf{\checkmark}} & \textcolor{robo_green}{\textbf{\checkmark}} & $16.27$ & $14.89$ & $23.54$ & $28.51$ & $29.98$
    \\\bottomrule
\end{tabular}}
\label{tab:ablation}
\end{table}

\begin{figure*}[t]
    \begin{center}
    \includegraphics[width=1.0\linewidth]{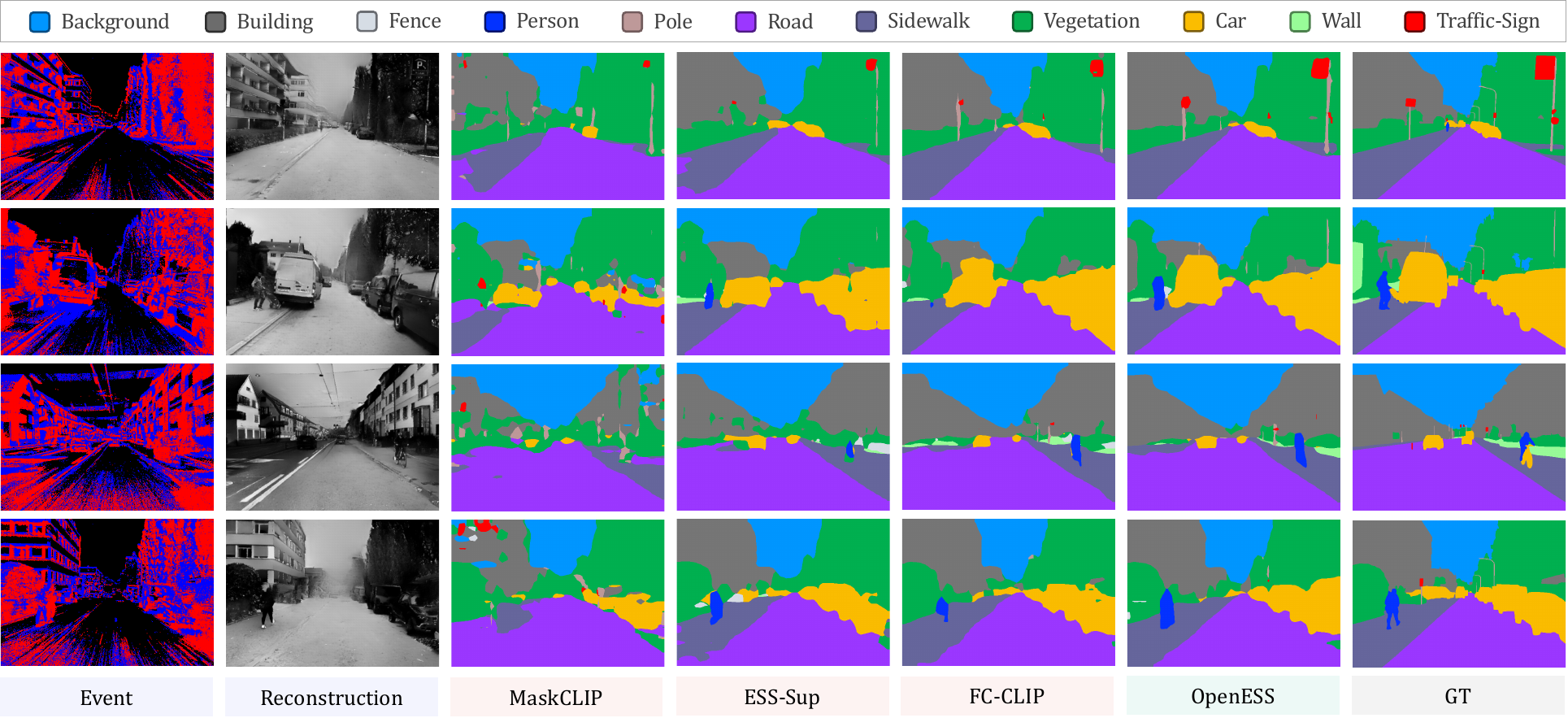}
    \end{center}
    \vspace{-0.4cm}
    \caption{\textbf{Qualitative comparisons} of state-of-the-art ESS approaches on the \textit{test} set of \textit{DSEC-Semantic} \cite{sun2022ess}. Each color corresponds to a distinct semantic category. GT denotes the ground truth semantic maps. Best viewed in colors and zoomed-in for additional details.}
    \label{fig:qualitative}
    \vspace{-0.15cm}
\end{figure*}

\noindent\textbf{Qualitative Assessment.}
\cref{fig:qualitative} provides visual comparisons between OpenESS and other approaches on \textit{DSEC-Semantic} \cite{sun2022ess}. We find that OpenESS tends to predict more consistent semantic information from sparse and irregular event inputs, especially at instance boundaries. We include more visual examples and failure cases in the Appendix.

\noindent\textbf{Open-World Predictions.}
One of the core advantages of OpenESS is the ability to predict beyond the fixed label set from the original training sets. As shown in \cref{fig:teaser}, our approach can take arbitrary text prompts as inputs and generate semantically coherent event predictions without using event labels. This is credited to the alignment between event features and CLIP's knowledge in T2E. Such a flexible way of prediction enables a more holistic event understanding.

\noindent\textbf{Other Representation Learning Approaches.}
In \cref{tab:pretrain}, we compare OpenESS with recent reconstruction-based \cite{he2022mae,bao2021beit,zhou2021ibot,yang2023ecdp} and contrastive learning-based \cite{chen2020simclr,chen2021mocov3} pre-training methods. As can be seen, the proposed OpenESS achieves competitive results over existing approaches. It is worth highlighting again that our framework distinct from prior arts by supporting open-vocabulary learning.

\subsection{Ablation Study}

\noindent\textbf{Cross-Modality Representation Learning.}
\cref{tab:ablation} provides a comprehensive ablation study on the frame-to-event (F2E) and text-to-event (T2E) learning objectives in OpenESS using three event representations. We observe that both F2E and T2E contribute to an overt improvement over random initialization under linear probing and few-shot fine-tuning settings, which verifies the effectiveness of our proposed approach. Once again, we find that the voxel grids tend to achieve better performance than other representations. The spike-based methods \cite{kim2022spiking}, albeit being computationally more efficient, show sub-par performance compared to voxel grids and reconstructions.

\begin{figure}[t]
    \begin{center}
    \includegraphics[width=1.0\linewidth]{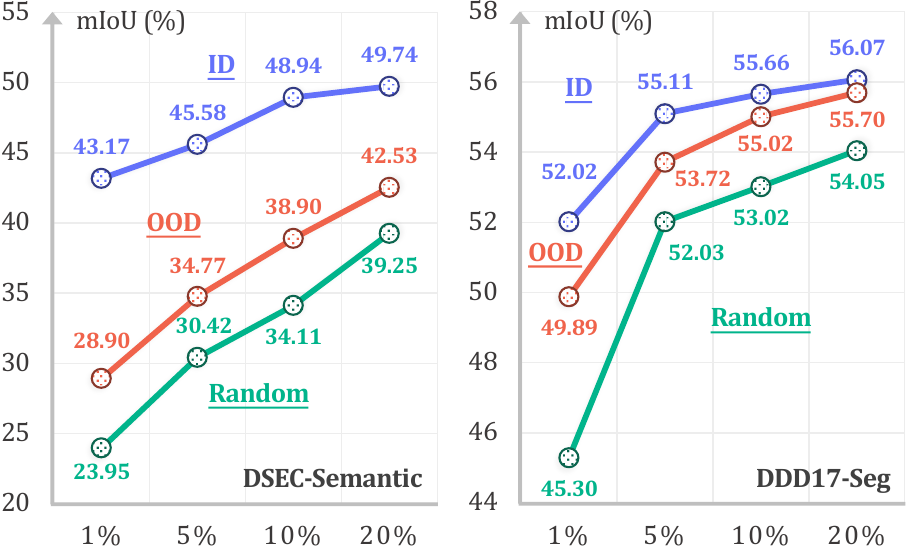}
    \end{center}
    \vspace{-0.4cm}
    \caption{\textbf{Cross-dataset representation learning} results of comparing OpenESS pre-training using in-distribution (ID) and out-of-distribution (OOD) data in-between the \textit{DDD17-Seg} \cite{binas2017ddd17} and \textit{DSEC-Semantic} \cite{sun2022ess} datasets. Models after pre-training are fine-tuned with 1\%, 5\%, 10\%, and 20\% annotations, respectively.}
    \label{fig:ood}
\end{figure}

\noindent\textbf{Superpixel Generation.}
We study the utilization of SLIC \cite{achanta2012slic} and SAM \cite{kirillov2023segment} in our frame-to-event contrastive distillation and show the results in \cref{fig:ablation}. Using either frame networks pre-trained by DINO \cite{caron2021dino}, MoCoV2 \cite{chen2020mocov2}, or SwAV \cite{caron2020swav}, the SAM-generated superpixels consistently exhibit better performance for event representation learning. The number of superpixels involved in calculating tends to affect the effectiveness of contrastive learning. A preliminary search to determine this hyperparameter is required.  We empirically find that setting $M$ to $100$ for \textit{DSEC-Semantic} \cite{sun2022ess} and $25$ for \textit{DDD17-Seg} \cite{alonso2019ev-segnet} will likely yield the best possible segmentation performance in our framework.

\noindent\textbf{Cross-Dataset Knowledge Transfer.}
Since we are targeting annotation-free representation learning, it is thus intuitive to see the cross-dataset adaptation effect. As shown in \cref{fig:ood}, pre-training on OOD datasets also brings appealing improvements over the random initialization baseline. This result highlights the importance of conducting representation learning for an effective transfer to downstream tasks. 

\begin{figure}[t]
    \begin{center}
    \includegraphics[width=1.0\linewidth]{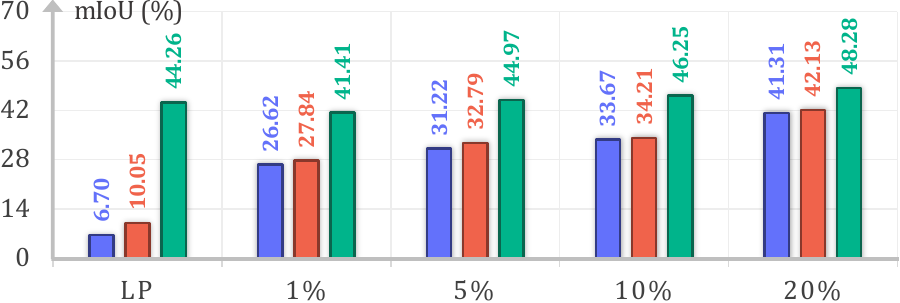}
    \end{center}
    \vspace{-0.45cm}
    \caption{\textbf{Single-modality OpenESS representation learning study} on the \textit{DSEC-Semantic} \cite{sun2022ess} dataset. The results are from models of random initialization (\colorsquare{robo_blue}), \texttt{recon2voxel} pre-training (\colorsquare{robo_red}), and \texttt{frame2voxel} pre-training (\colorsquare{robo_green}), respectively, after linear probing (LP) and annotation-efficient fine-tuning.}
    \label{fig:recon}
\end{figure}
\noindent\textbf{Framework with Event Camera Only.}
Lastly, we study the scenario where the frame camera becomes unavailable. We replace the input to the frame branch with event reconstructions \cite{rebecq2019e2vid} and show the results in \cref{fig:recon}. Since the limited visual cues from the reconstruction tend to degrade the quality of representation learning, its performance is sub-par compared to the frame-based knowledge transfer.

\vspace{0.125cm}
\section{Conclusion}
\label{sec:conclusion}
In this work, we introduced OpenESS, an open-vocabulary event-based semantic segmentation framework tailored to perform open-vocabulary ESS in an annotation-efficient manner. We proposed to encourage cross-modality representation learning between events and frames using frame-to-event contrastive distillation and text-to-event semantic consistency regularization. Through extensive experiments, we validated the effectiveness of OpenESS in tackling dense event-based predictions. We hope this work could shed light on the future development of more scalable ESS systems.

\vspace{0.1cm}
{\small
\noindent\textbf{Acknowledgement.}
This work is under the programme DesCartes and is supported by the National Research Foundation, Prime Minister’s Office, Singapore, under its Campus for Research Excellence and Technological Enterprise (CREATE) programme.
}

\clearpage
\appendix
\section*{Appendix}
\startcontents[appendices]
\printcontents[appendices]{l}{1}{\setcounter{tocdepth}{3}}

\section{Additional Implementation Details}
In this section, we provide additional details to assist the implementation and reproduction of the approaches in the proposed OpenESS framework.

\subsection{Datasets}
In this study, we follow prior works \cite{sun2022ess,jia2023evsegformer,alonso2019ev-segnet,hamaguchi2023hmnet} by using the \textit{\textbf{DDD17-Seg}} \cite{alonso2019ev-segnet} and \textit{\textbf{DSEC-Semantic}} \cite{sun2022ess} datasets for evaluating and validating the baselines, prior methods, and the proposed OpenESS framework. Some specifications related to these two datasets are listed as follows.

\begin{table*}[t]
    \centering
    \caption{The specifications of the \textit{DDD17-Seg} dataset \cite{alonso2019ev-segnet}.}
    \vspace{-0.1cm}
    \scalebox{0.95}{\begin{tabular}{l|c|c|c|c|c|c}
    \toprule
    \textbf{-} & \multicolumn{5}{c|}{\textbf{Training}} & \textbf{Test}
    \\\midrule\midrule
    \textbf{Seq} & \texttt{dir0} & \texttt{dir3} & \texttt{dir4} & \texttt{dir6} & \texttt{dir7} & \texttt{dir1}
    \\\midrule
    \# Frames & $11785$ & $20051$ & $41071$ & $28411$ & $58650$ & $71680$
    \\\midrule
    \# Events & $5550$ & $1320$ & $6945$ & $1140$ & $995$ & $3890$
    \\\midrule
    Resolution & \multicolumn{5}{c|}{$352\times 200$} & $352\times 200$
    \\\midrule
    \# Classes & \multicolumn{5}{c|}{$6$ Classes} & $6$ Classes
    \\\bottomrule
    \end{tabular}}
    \label{tab:dataset_ddd17}
    \vspace{0.2cm}
\end{table*}
\begin{table*}[t]
    \centering
    \caption{The specifications of the \textit{DSEC-Semantic} dataset \cite{sun2022ess}.}
    \vspace{-0.1cm}
    \scalebox{0.95}{\begin{tabular}{l|c|c|c|c|c|c|c|c|c|c|c}
    \toprule
    \textbf{-} & \multicolumn{8}{c|}{\textbf{Training}} & \multicolumn{3}{c}{\textbf{Test}}
    \\\midrule\midrule
    \textbf{Seq} & \texttt{00\_a} & \texttt{01\_a} & \texttt{02\_a} & \texttt{04\_a} & \texttt{05\_a} & \texttt{06\_a} & \texttt{07\_a} & \texttt{08\_a} & \texttt{13\_a} & \texttt{14\_c} & \texttt{15\_a}
    \\\midrule
    \# Frames & $939$ & $681$ & $235$ & $701$ & $1753$ & $1523$ & $1463$ & $787$ & $379$ & $1191$ & $1239$
    \\\midrule
    \# Events & $933$ & $675$ & $229$ & $695$ & $1747$ & $1517$ & $1457$ & $781$ & $373$ & $1185$ & $1233$
    \\\midrule
    Resolution & \multicolumn{8}{c|}{$640\times 440$} & \multicolumn{3}{c}{$640\times 440$}
    \\\midrule
    \# Classes & \multicolumn{8}{c|}{$11$ Classes} & \multicolumn{3}{c}{$11$ Classes}
    \\\bottomrule
    \end{tabular}}
    \label{tab:dataset_dsec11}
\end{table*}

\begin{itemize}
    \item \textbf{DDD17-Seg} \cite{alonso2019ev-segnet} serves as the first benchmark for ESS. It is a semantic segmentation extension of the DDD17 \cite{binas2017ddd17} dataset, which includes hours of driving data, capturing a variety of driving conditions such as different times of day, traffic scenarios, and weather conditions. Alonso and Murillo \cite{alonso2019ev-segnet} provide the semantic labels on top of DDD17 to enable event-based semantic segmentation. Specifically, they proposed to use the corresponding gray-scale images along with the event streams to generate an approximated set of semantic labels for training, which was proven effective in training models to segment directly on event-based data. A three-step procedure is applied: \textit{i)} train a semantic segmentation model on the gray-scale images in the \textit{Cityscapes} dataset \cite{cordts2016cityscapes}; \textit{ii)} Use the trained model to label the gray-scale images in DDD17; and \textit{iii)} Conduct a post-processing step on the generated pseudo labels, including class merging and image cropping. The dataset specification is shown in \cref{tab:dataset_ddd17}. In total, there are $15950$ training and $3890$ test samples in the DDD17-Seg dataset. Each pixel is labeled across six semantic classes, including \texttt{flat}, \texttt{background}, \texttt{object}, \texttt{vegetation}, \texttt{human}, and \texttt{vehicle}. For each sample, we convert the event streams into a sequence of $20$ voxel grids, each consisting of $32000$ events and with a spatial resolution of $352\times 200$. For additional details of this dataset, kindly refer to \url{http://sensors.ini.uzh.ch/news_page/DDD17.html}.

    \item \textbf{DSEC-Semantic} \cite{sun2022ess} is a semantic segmentation extension of the DSEC (Driving Stereo Event Camera) dataset \cite{gehrig2021dsec}. DSEC is an extensive dataset designed for advanced driver-assistance systems (ADAS) and autonomous driving research, with a particular focus on event-based vision and stereo vision. Different from DDD17 \cite{binas2017ddd17}, the DSEC dataset combines data from event-based cameras and traditional RGB cameras. The inclusion of event-based cameras (which capture changes in light intensity) alongside regular cameras provides a rich, complementary data source for perception tasks. The dataset typically features high-resolution images and event data, providing detailed visual information from a wide range of driving conditions, including urban, suburban, and highway environments, various weather conditions, and different times of the day. This diversity is crucial for developing systems that can operate reliably in real-world conditions. Based on such a rich collection, Sun \etal \cite{sun2022ess} adopted a similar pseudo labeling procedure as DDD17-Seg \cite{alonso2019ev-segnet} and generated the semantic labels for eleven sequences in DSEC, dubbed as DSEC-Semantic. The dataset specification is shown in \cref{tab:dataset_dsec11}. In total, there are $8082$ training and $2809$ test samples in the DSEC-Semantic dataset. Each pixel is labeled across eleven semantic classes, including \texttt{background}, \texttt{building}, \texttt{fence}, \texttt{person}, \texttt{pole}, \texttt{road}, \texttt{sidewalk}, \texttt{vegetation}, \texttt{car}, \texttt{wall}, and \texttt{traffic-sign}. For each sample, we convert the event streams into a sequence of $20$ voxel grids, each consisting of $100000$ events and with a spatial resolution of $640\times 440$. For additional details of this dataset, kindly refer to \url{https://dsec.ifi.uzh.ch/dsec-semantic}.
\end{itemize}

\subsection{Text Prompts}
To enable the conventional evaluation of our proposed open-vocabulary approach on an event-based semantic segmentation dataset, we need to use the pre-defined class names as text prompts to generate the text embedding. Specifically, we follow the standard templates \cite{radford2021clip} when generating the embedding. The dataset-specific text prompts defined in our framework are listed as follows.

\begin{table*}[t]
    \centering
    \caption{The text prompts defined on the \textit{DDD17-Seg} dataset \cite{alonso2019ev-segnet} ($6$ classes) used for generating the CLIP text embedding.}
    \vspace{-0.1cm}
    \scalebox{0.95}{\begin{tabular}{c|l|p{11.5cm}}
    \toprule
    \multicolumn{3}{c}{\textbf{DDD17 (6 classes)}}
    \\\midrule\midrule
    \textbf{\#} & \textbf{class} & \textbf{text prompt}
    \\\midrule
    $0$ & \texttt{flat} & `road', `driveable', `street', `lane marking', `bicycle lane', `roundabout lane', `parking lane', `terrain', `grass', `soil', `sand', `lawn', `meadow', `turf'
    \\\midrule
    $1$ & \texttt{background} & `sky', `building'
    \\\midrule
    $2$ & \texttt{object} & `pole', `traffic sign pole', `traffic light pole', `traffic light box', `traffic-sign', `parking-sign', `direction-sign'
    \\\midrule
    $3$ & \texttt{vegetation} & `vegetation', `vertical vegetation', `tree', `tree trunk', `hedge', `woods', `terrain', `grass', `soil', `sand', `lawn', `meadow', `turf'
    \\\midrule
    $4$ & \texttt{human} & `person', `pedestrian', `walking people', `standing people', `sitting people', `toddler'
    \\\midrule
    $5$ & \texttt{vehicle} & `car', `jeep', `SUV', `van', `caravan', `truck', `box truck', `pickup truck', `trailer', `bus', `public bus', `train', `vehicle-on-rail', `tram', `motorbike', `moped', `scooter', `bicycle'
    \\\bottomrule
    \end{tabular}}
    \label{tab:prompt_ddd17}
    \vspace{0.2cm}
\end{table*}
\begin{table*}[t]
    \centering
    \caption{The ext prompts defined on the \textit{DSEC-Semantic} dataset \cite{sun2022ess} ($11$ classes) used for generating the CLIP text embedding.}
    \vspace{-0.1cm}
    \scalebox{0.95}{\begin{tabular}{c|l|p{11.5cm}}
    \toprule
    \multicolumn{3}{c}{\textbf{DSEC-Semantic (11 classes)}}
    \\\midrule\midrule
    \textbf{\#} & \textbf{class} & \textbf{text prompt}
    \\\midrule
    $0$ & \texttt{background} & `sky'
    \\\midrule
    $1$ & \texttt{building} & `building', `skyscraper', `house', `bus stop building', `garage', `carport', `scaffolding'
    \\\midrule
    $2$ & \texttt{fence} & `fence', `fence with hole'
    \\\midrule
    $3$ & \texttt{person} & `person', `pedestrian', `walking people', `standing people', `sitting people', `toddler'
    \\\midrule
    $4$ & \texttt{pole} & `pole', `electric pole', `traffic sign pole', `traffic light pole'
    \\\midrule
    $5$ & \texttt{road} & `road', `driveable', `street', `lane marking', `bicycle lane', `roundabout lane', `parking lane'
    \\\midrule
    $6$ & \texttt{sidewalk} & `sidewalk', `delimiting curb', `traffic island', `walkable', `pedestrian zone'
    \\\midrule
    $7$ & \texttt{vegetation} & `vegetation', `vertical vegetation', `tree', `tree trunk', `hedge', `woods', `terrain', `grass', `soil', `sand', `lawn', `meadow', `turf'
    \\\midrule
    $8$ & \texttt{car} & `car', `jeep', `SUV', `van', `caravan', `truck', `box truck', `pickup truck', `trailer', `bus', `public bus', `train', `vehicle-on-rail', `tram', `motorbike', `moped', `scooter', `bicycle'
    \\\midrule
    $9$ & \texttt{wall} & `wall', `standing wall'
    \\\midrule
    $10$ & \texttt{traffic-sign} & `traffic-sign', `parking-sign', `direction-sign', `traffic-sign without pole', `traffic light box'
    \\\bottomrule
    \end{tabular}}
    \label{tab:prompt_dsec11}
\end{table*}

\begin{itemize}
    \item \textbf{DDD17-Seg.} There is a total of six semantic classes in the DDD17-Seg dataset \cite{alonso2019ev-segnet}, with static and dynamic components of driving scenes. Our defined text prompts of this dataset are summarized in \cref{tab:prompt_ddd17}. For each semantic class, we generate for each text prompt the text embedding using the CLIP text encoder and then average the text embedding of all text prompts as the final embedding of this class.

    \item \textbf{DSEC-Semantic.} There is a total of eleven semantic classes in the DSEC-Semantic dataset \cite{sun2022ess}, ranging from static and dynamic components of driving scenes. Our defined text prompts of this dataset are summarized in \cref{tab:prompt_dsec11}. For each semantic class, we generate for each text prompt the text embedding using the CLIP text encoder and then average the text embedding of all text prompts as the final embedding of this class.
\end{itemize}

\begin{figure*}[t]
    \begin{center}
    \includegraphics[width=1.0\linewidth]{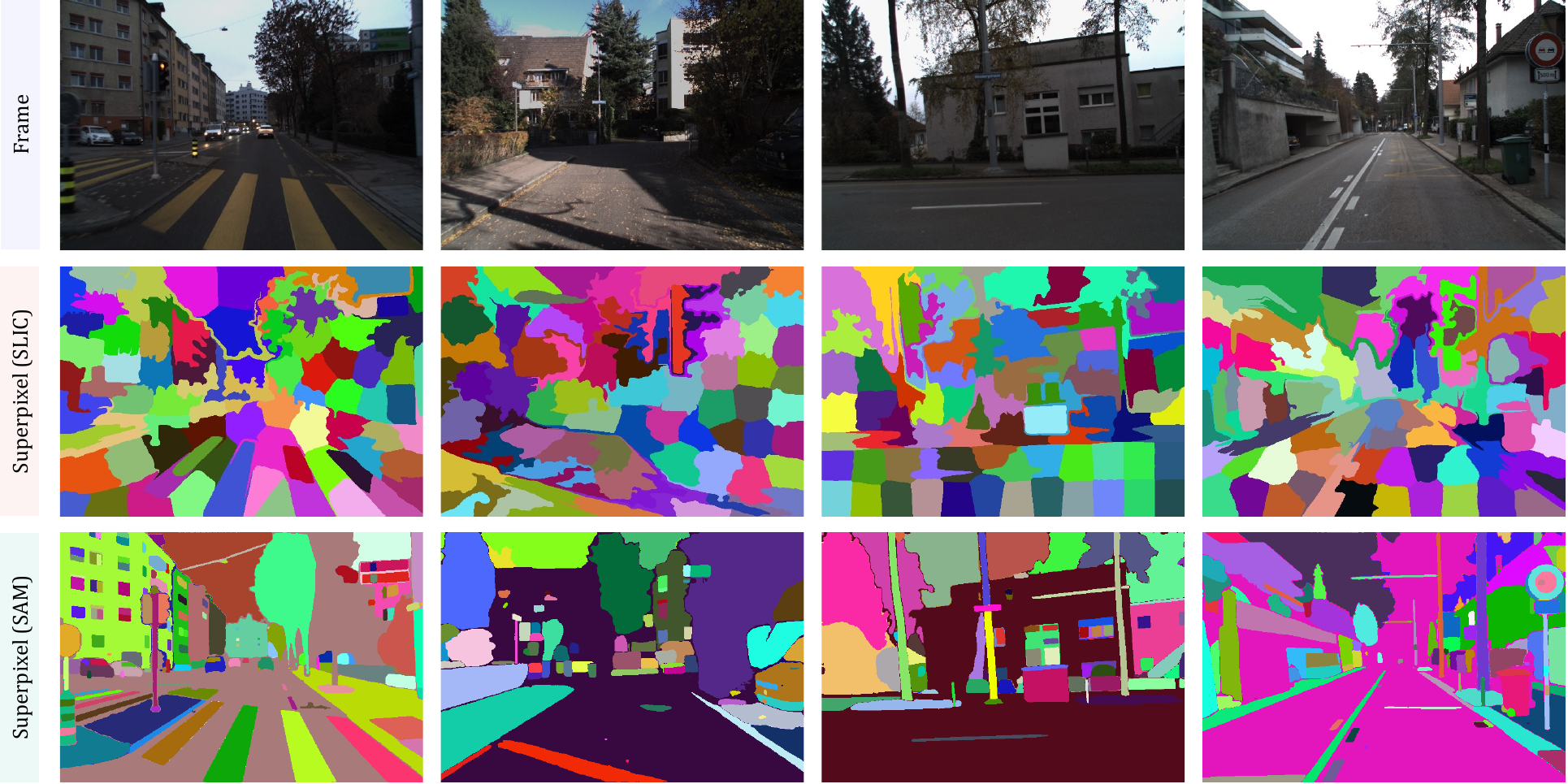}
    \end{center}
    \vspace{-0.45cm}
    \caption{\textbf{Examples of superpixels} generated by SLIC \cite{achanta2012slic} (the $2$nd row) and SAM \cite{kirillov2023segment} (the $3$rd row). The parameter $M_{slic}$ in the SLIC algorithm is set to $100$. Each colored patch represents one distinct and semantically coherent superpixel. Best viewed in colors.}
    \label{fig:superpixel_slic_sam}
\end{figure*}

\subsection{Superpixels}
In image processing and computer vision, superpixels can be defined as a scheme that groups pixels in an image into perceptually meaningful atomic regions, which are used to replace the rigid structure of the pixel grid \cite{achanta2012slic}. Superpixels provide a more natural representation of the image structure, often leading to more efficient and effective image processing. Here are some of their key aspects:
\begin{itemize}
    \item \textbf{Grouping Pixels.} Superpixels are often formed by clustering pixels based on certain criteria like color similarity, brightness, texture, and other low-level patterns \cite{achanta2012slic}, or more recently, semantics \cite{kirillov2023segment}. This results in contiguous regions in the image that are more meaningful than individual pixels for many applications \cite{liu2023seal,chen2023towards,peng2023learning,xu2024visual}.

    \item \textbf{Reducing Complexity.} By aggregating pixels into superpixels, the complexity of image data is significantly reduced \cite{stutz2018superpixels}. This reduction helps in speeding up subsequent image processing tasks, as algorithms have fewer elements (superpixels) to process compared to the potentially millions of pixels in an image.

    \item \textbf{Preserving Edges.} One of the primary goals of superpixel segmentation is to preserve important image edges. Superpixels often adhere closely to the boundaries of objects in the image, making them useful for tasks that rely on accurate edge information, like object recognition and scene understanding.
\end{itemize}

In this work, we propose to first leverage calibrated frames to generate coarse, instance-level superpixels and then distill knowledge from a pre-trained image backbone to the event segmentation network. Specifically, we resort to the following two ways to generate the superpixels. 

\begin{itemize}
    \item \textbf{SLIC.} The first way is to leverage the heuristic Simple Linear Iterative Clustering (SLIC) approach \cite{achanta2012slic} to efficiently group pixels from frame $I^{img}_i$ into a total of $M_{slic}$ segments with good boundary adherence and regularity. The superpixels are defined as $I^{sp}_i=\{\mathcal{I}_i^1, \mathcal{I}_i^2, ..., \mathcal{I}_i^{M_{slic}}\}$, where $M_{slic}$ is a hyperparameter that needs to be adjusted based on the inputs. The generated superpixels satisfy $\mathcal{I}_i^1\cup \mathcal{I}_i^2 \cup ... \cup \mathcal{I}_i^{M_{slic}} = \{ 1, 2, ..., H\times W\}$. Several examples of the SLIC-generated superpixels are shown in the second row of \cref{fig:superpixel_slic_sam}, where each of the color-coded patches represents one distinct and semantically coherent superpixel.

    \item \textbf{SAM.} For the second option, we use the recent Segment Anything Model (SAM) \cite{kirillov2023segment} which takes $I^{img}_i$ as the input and outputs $M_{sam}$ class-agnostic masks. For simplicity, we use $M$ to denote the number of superpixels used during knowledge distillation, \ie, $\{I^{sp}_i=\{\mathcal{I}_i^1,...,\mathcal{I}_i^k\}|k=1,...,M\}$. Several examples of the SAM-generated superpixels are shown in the third row of \cref{fig:superpixel_slic_sam}, where each of the color-coded patches represents one distinct and semantically coherent superpixel.
    
\end{itemize}

\begin{figure*}[t]
    \vspace{-0.37cm}
    \centering
    \begin{subfigure}[b]{.49\textwidth}
         \centering
         \includegraphics[width=\textwidth]{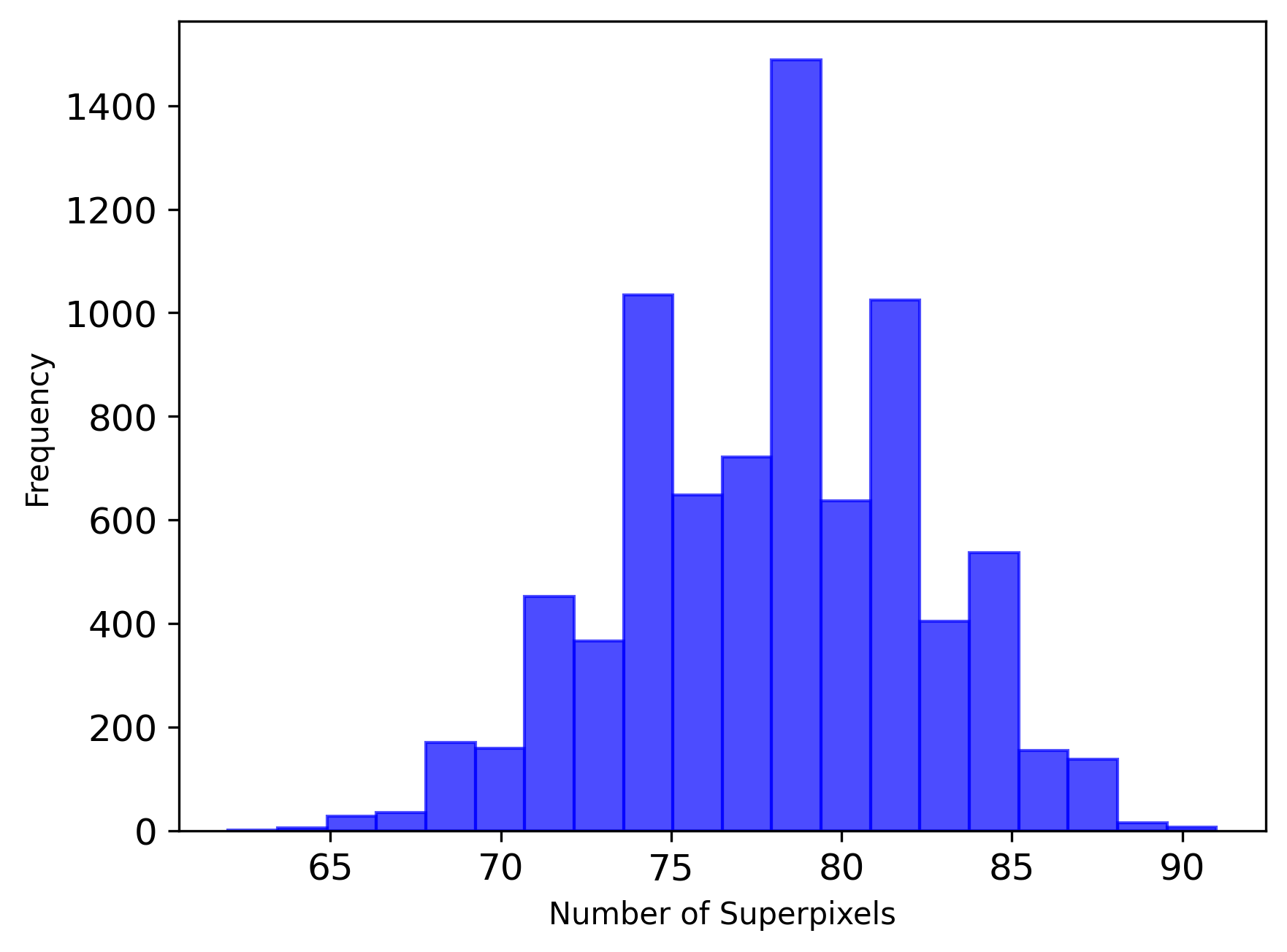}
         \caption{Histogram of SLIC-Generated Superpixels}
         \label{fig:histogram_slic}
    \end{subfigure}
    \hfill
    \begin{subfigure}[b]{.49\textwidth}
         \centering
         \includegraphics[width=\textwidth]{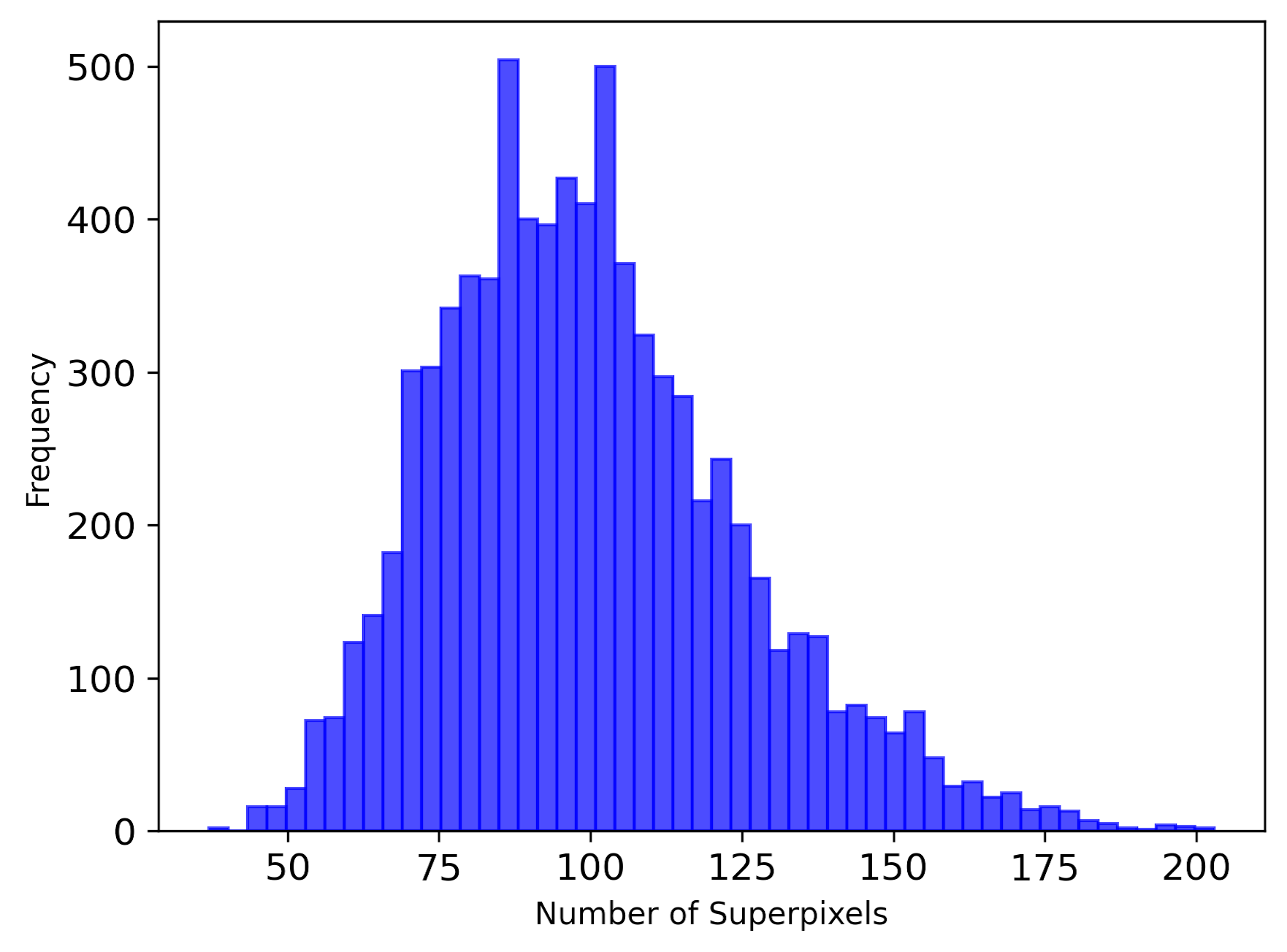}
         \caption{Histogram of SAM-Generated Superpixels}
         \label{fig:histogram_sam}
    \end{subfigure}
    \caption{\textbf{The statistical distributions} of superpixels generated by SLIC \cite{achanta2012slic} (subfigure a) and SAM \cite{kirillov2023segment} (subfigure b).}
    \label{fig:superpixel_histograms}
\end{figure*}

\begin{figure*}[t]
    \begin{center}
    \includegraphics[width=1.0\linewidth]{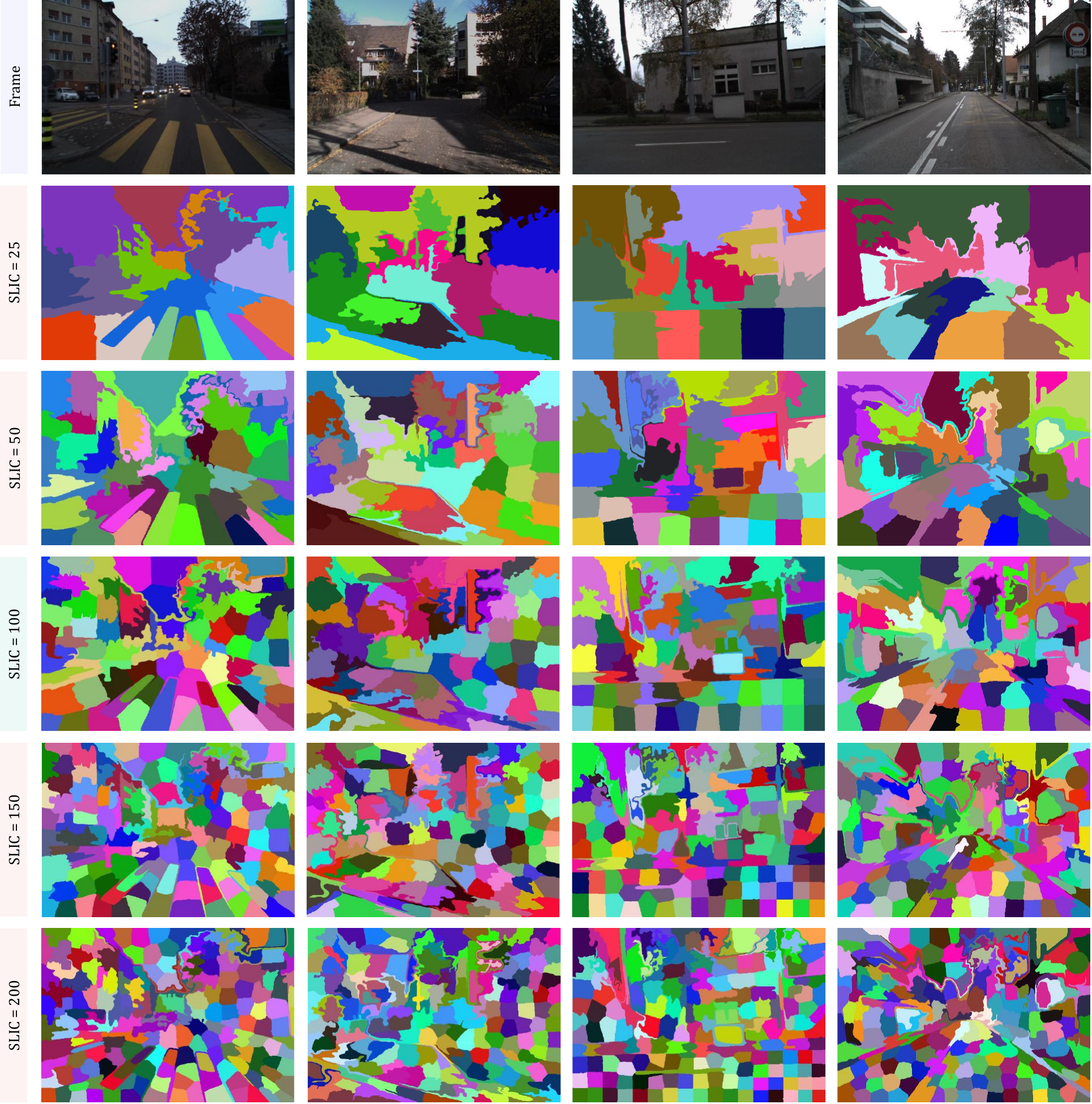}
    \end{center}
    \vspace{-0.45cm}
    \caption{\textbf{Examples of superpixels} generated by SLIC \cite{achanta2012slic} with different numbers of superpixels $M_{slic}$ ($25$, $50$, $100$, $150$, and $200$). Each colored patch represents one distinct and semantically coherent superpixel. Best viewed in colors.}
    \label{fig:superpixel_slic}
\end{figure*}

We calculate the SLIC and SAM superpixel distributions on the training set of the DSEC-Semantic dataset \cite{sun2022ess} and show the corresponding statistics in \cref{fig:superpixel_histograms}. As can be observed, the SLIC-generated superpixels often contain more low-level visual cues, such as color similarity, brightness, and texture. On the contrary, superpixels generated by SAM exhibit clear semantic coherence and often depict the boundaries of objects and backgrounds. As verified in the main body of this paper, the semantically richer SAM superpixels bring higher performance gains in our Frame-to-Event Contrastive Learning framework.

Meanwhile, we provide more fine-grained examples of the SLIC algorithm using different $M_{slic}$, \ie, $25$, $50$, $100$, $150$, and $200$. The results are shown in \cref{fig:superpixel_slic}. Specifically, the number of superpixels $M_{slic}$ should reflect the complexity and detail of the image. For images with high detail or complexity (like those with many objects or textures), a larger $M_{slic}$ can capture more of this detail. Conversely, for simpler images, fewer superpixels might be sufficient. Usually, more superpixels mean smaller superpixels. Smaller superpixels can adhere more closely to object boundaries and capture finer details, but they might also capture more noise. Fewer superpixels result in larger, more homogeneous regions but may lead to a loss of detail, especially at the edges of objects. The choice also depends on the specific application. For instance, in object detection or segmentation tasks where boundary adherence is crucial, a higher number of superpixels might be preferable. In contrast, for tasks like image compression or abstraction, fewer superpixels might be more appropriate. Often, the optimal number of superpixels is determined empirically. This involves experimenting with different values and evaluating the results based on the specific criteria of the task or application. In our event-based semantic segmentation task, we choose $M_{slic}=100$ for our Frame-to-Event Contrastive Learning on the DSEC-Semantic dataset \cite{sun2022ess}, and $M_{slic}=25$ on the DDD17-Seg dataset \cite{alonso2019ev-segnet}.

Since $I^{evt}_i$ and $I^{img}_i$ have been aligned and synchronized, we can group events from $I^{evt}_i$ into superevents $\{V^{sp}_i=\{\mathcal{V}_i^1,...,\mathcal{V}_i^l\}|l=1,...,M\}$ by using the known event-pixel correspondences.

\begin{table*}[t]
    \centering
    \caption{\textbf{The per-class segmentation results} of annotation-free event-based semantic segmentation approaches on the test set of \textit{DSEC-Semantic} \cite{sun2022ess}. Scores reported are IoUs in percentage (\%). For each semantic class, the best score in each column is highlighted in \textbf{bold}.}
    \vspace{-0.1cm}
    \scalebox{0.89}{\begin{tabular}{r|c|ccccccccccc|c}
    \toprule
    \textbf{Method} & \rotatebox{0}{\textbf{mIoU}} & \rotatebox{90}{background~} & \rotatebox{90}{building} & \rotatebox{90}{fence} & \rotatebox{90}{person} & \rotatebox{90}{pole} & \rotatebox{90}{road} & \rotatebox{90}{sidewalk} & \rotatebox{90}{vegetation} & \rotatebox{90}{car} & \rotatebox{90}{wall} & \rotatebox{90}{traffic-sign} & \textbf{Acc}
    \\\midrule\midrule
    \rowcolor{robo_green!7}\multicolumn{14}{l}{\textcolor{robo_green}{\textbf{Annotation-Free ESS}}}
    \\
    MaskCLIP \cite{zhou2022extract} & $21.97$ & $26.45$ & $52.59$ & $0.20$ & $0.04$ & $\mathbf{4.19}$ & $65.76$ & $2.96$ & $48.02$ & $40.67$ & $0.67$ & $0.08$ & $58.96$
    \\
    FC-CLIP \cite{yu2023fc} & $39.42$ & $87.49$ & $69.68$ & $\mathbf{14.39}$ & $\mathbf{17.53}$ & $0.29$ & $71.76$ & $34.56$ & $71.30$ & $63.19$ & $2.98$ & $0.50$ & $79.20$
    \\
    \textbf{OpenESS (Ours)} & $\mathbf{43.31}$ & $\mathbf{92.53}$ & $\mathbf{74.22}$ & $11.96$ & $0.00$ & $0.41$ & $\mathbf{87.32}$ & $\mathbf{55.09}$ & $\mathbf{74.23}$ & $\mathbf{64.25}$ & $\mathbf{7.98}$ & $\mathbf{8.47}$ & $\mathbf{86.18}$
    \\\bottomrule
    \end{tabular}}
    \label{tab:per_class_dsec11}
\end{table*}

\subsection{Backbones}
As mentioned in the main body of this paper, we establish three open-vocabulary event-based semantic segmentation settings based on the use of three different event representations, \ie, \texttt{frame2voxel}, \texttt{frame2recon}, and \texttt{frame2spike}. It is worth noting that these three event representations tend to have their own advantages.

We supplement additional implementation details regarding the used event representations as follows.

\begin{itemize}

    \item \textbf{Frame2Voxel.} For the use of \textit{voxel grids} as the event embedding, we follow Sun \etal \cite{sun2022ess} by converting the raw events $\varepsilon_i$ into the regular voxel grids $I^{vox}_i\in\mathbb{R}^{C\times H\times W}$ as the input to the event-based semantic segmentation network. This representation is intuitive and aligns well with conventional event camera data processing techniques. It is suitable for convolutional neural networks as it maintains spatial and temporal relationships. Specifically, with a predefined number of events, each voxel grid is built from non-overlapping windows as follows:
    
    \begin{equation}
    I^{vox}_i = \sum_{\mathbf{e}_j\in \varepsilon_i} p_j \delta(\mathbf{x}_j - \mathbf{x}) \delta(\mathbf{y}_j - \mathbf{y}) \max\{1 - |t^{*}_j - t| , 0\},~
    \label{eq:vox}
    \end{equation}
    
    where $\delta$ is the Kronecker delta function; $t^{*}_j = (B-1)\frac{t_j - t_0}{\Delta T}$ is the normalized event timestamp with $B$ as the number of temporal bins in an event stream; $\Delta T$ is the time window and $t_0$ denotes the time of the first event in the window. It is worth noting that \textit{voxel grids} can be memory-intensive, especially for high-resolution sensors or long-time windows. They might also introduce quantization errors due to the discretization of space and time.
    For additional details on the use of \textit{voxel grids}, kindly refer to \url{https://github.com/uzh-rpg/ess}.

    \item \textbf{Frame2Recon.} For the use of \textit{event reconstructions} as the event embedding, we follow Sun \etal \cite{sun2022ess} and Rebecq \etal \cite{rebecq2019e2vid} by converting the raw events $\varepsilon_i$ into the regular frame-like event reconstructions $I^{rec}_i\in\mathbb{R}^{H\times W}$ as the input to the event-based semantic segmentation network. This can be done by accumulating events over short time intervals or by using algorithms to interpolate or simulate frames. This approach is compatible with standard image processing techniques and algorithms developed for frame-based vision. It is more familiar to practitioners used to working with conventional cameras. In this work, we adopt the E2VID model \cite{rebecq2019e2vid} to generate the \textit{event reconstructions}. This process can be described as follows:
    
    \begin{align}
    \mathbf{z}^{rec}_k =&~ ~E_{\text{e2vid}} (I^{vox}_k, \mathbf{z}^{rec}_{k-1}), ~~~ k = 1, ..., N,~
    \\
    I^{rec}_i =&~ ~D_{\text{e2vid}} (\mathbf{z}^{rec}),~
    \end{align}
    where $I^{vox}_k$ denotes the \textit{voxel grids} as defined in \cref{eq:vox}; $E_{\text{e2vid}}$ and $D_{\text{e2vid}}$ are the encoder of decoder of the E2VID model \cite{rebecq2019e2vid}, respectively. It is worth noting that \textit{event reconstructions} can lose the fine temporal resolution that event cameras provide. They might also introduce artifacts or noise, especially in scenes with fast-moving objects or low event rates. 
    For additional details on the use of \textit{event reconstructions}, kindly refer to \url{https://github.com/uzh-rpg/rpg_e2vid}.

    \item \textbf{Frame2Spike.} For the use of \textit{spikes} as the event embedding, we follow Kim \etal \cite{kim2022spiking} by converting the raw events $\varepsilon_i$ into spikes $I^{spk}_i\in\mathbb{R}^{H\times W}$ as the input to the event-based semantic segmentation network. The spike representation keeps the data in its raw form – as individual spikes or events.  This representation preserves the high temporal resolution of the event data and is highly efficient in terms of memory and computation, especially for sparse scenes. The rate coding is used as the spike encoding scheme due to its reliable performance across various tasks. Each pixel value with a random number ranging between $[s_{min}, s_{max}]$ at every time step is recorded, where $s_{min}$ and $s_{max}$ are the minimum and maximum possible pixel intensities, respectively. If the random number is greater than the pixel intensity, the Poisson spike generator outputs a spike with amplitude $1$. Otherwise, the Poisson spike generator does not yield any spikes. The spikes in a certain time window are accumulated to generate a frame, where such frames will serve as the input to the event-based semantic segmentation network. It is worth noting that processing raw spike data requires specialized algorithms, often inspired by neuromorphic computing. It might not be suitable for traditional image processing techniques and can be challenging to interpret and visualize.
    For additional details on the use of \textit{spikes}, kindly refer to \url{https://github.com/Intelligent-Computing-Lab-Yale/SNN-Segmentation}.
    
\end{itemize}

To sum up, each event representation has its unique characteristics and is suitable for different applications or processing techniques. Our proposed OpenESS framework is capable of leveraging each of the above event representations for efficient and accurate event-based semantic segmentation in an annotation-free and open-vocabulary manner. Such a versatile and flexible way of learning verifies the broader application potential of our proposed framework.

\begin{table*}[t]
    \centering
    \caption{\textbf{The per-class segmentation results} of annotation-efficient event-based semantic segmentation approaches on the test set of \textit{DSEC-Semantic} \cite{sun2022ess}. All approaches adopted the \texttt{frame2voxel} representation. Scores reported are IoUs in percentage (\%). For each semantic class under each experimental setting, the best score in each column is highlighted in \textbf{bold}.}
    \vspace{-0.1cm}
    \scalebox{0.865}{\begin{tabular}{r|c|ccccccccccc|c}
    \toprule
    \textbf{Method} & \rotatebox{90}{\textbf{mIoU}} & \rotatebox{90}{background~} & \rotatebox{90}{building} & \rotatebox{90}{fence} & \rotatebox{90}{person} & \rotatebox{90}{pole} & \rotatebox{90}{road} & \rotatebox{90}{sidewalk} & \rotatebox{90}{vegetation} & \rotatebox{90}{car} & \rotatebox{90}{wall} & \rotatebox{90}{traffic-sign} & \textbf{Acc}
    \\\midrule\midrule
    \rowcolor{robo_blue!7}\multicolumn{14}{l}{\textcolor{robo_blue}{\textbf{Linear Probing}}}
    \\
    \textcolor{gray}{Random} & \textcolor{gray}{$6.70$} & \textcolor{gray}{$7.85$} & \textcolor{gray}{$3.37$} & \textcolor{gray}{$0.00$} & \textcolor{gray}{$0.00$} & \textcolor{gray}{$0.00$} & \textcolor{gray}{$38.60$} & \textcolor{gray}{$0.00$} & \textcolor{gray}{$23.83$} & \textcolor{gray}{$0.01$} & \textcolor{gray}{$0.00$} & \textcolor{gray}{$0.00$} & \textcolor{gray}{$37.94$}
    \\
    MaskCLIP \cite{zhou2022extract} & $33.08$ & $75.04$ & $65.06$ & $4.63$ & $0.00$ & $\mathbf{6.47}$ & $77.06$ & $17.07$ & $55.89$ & $52.17$ & $0.69$ & $\mathbf{9.78}$ & $76.39$
    \\
    FC-CLIP \cite{yu2023fc} & $43.00$ & $92.53$ & $72.59$ & $\mathbf{12.43}$ & $0.02$ & $0.00$ & $88.14$ & $52.84$ & $71.92$ & $64.02$ & $\mathbf{10.54}$ & $7.95$ & $86.00$
    \\
    \textbf{OpenESS (Ours)} & $\mathbf{44.26}$ & $\mathbf{93.64}$ & $\mathbf{75.40}$ & $11.82$ & $\mathbf{1.16}$ & $0.75$ & $\mathbf{90.29}$ & $\mathbf{57.96}$ & $\mathbf{73.15}$ & $\mathbf{65.36}$ & $9.69$ & $7.67$ & $\mathbf{87.55}$
    \\\midrule
    \rowcolor{robo_red!7}\multicolumn{14}{l}{\textcolor{robo_red}{\textbf{Fine-Tuning (1\%)}}}
    \\
    \textcolor{gray}{Random} & \textcolor{gray}{$26.62$} & \textcolor{gray}{$81.63$} & \textcolor{gray}{$33.13$} & \textcolor{gray}{$1.77$} & \textcolor{gray}{$0.97$} & \textcolor{gray}{$7.58$} & \textcolor{gray}{$76.81$} & \textcolor{gray}{$17.45$} & \textcolor{gray}{$51.05$} & \textcolor{gray}{$18.64$} & \textcolor{gray}{$0.37$} & \textcolor{gray}{$3.40$} & \textcolor{gray}{$70.04$}
    \\
    MaskCLIP \cite{zhou2022extract} & $33.89$ & $87.56$ & $53.24$ & $2.34$ & $0.60$ & $8.92$ & $81.71$ & $25.76$ & $59.37$ & $42.56$ & $2.52$ & $8.24$ & $77.79$
    \\
    FC-CLIP \cite{yu2023fc} & $39.12$ & $91.64$ & $59.78$ & $\mathbf{8.93}$ & $0.00$ & $7.84$ & $\mathbf{87.58}$ & $\mathbf{46.58}$ & $66.87$ & $51.30$ & $\mathbf{4.74}$ & $5.10$ & $82.12$
    \\
    \textbf{OpenESS (Ours)} & $\mathbf{41.41}$ & $\mathbf{93.01}$ & $\mathbf{74.01}$ & $3.21$ & $\mathbf{10.78}$ & $\mathbf{14.58}$ & $84.50$ & $34.78$ & $\mathbf{69.82}$ & $\mathbf{55.12}$ & $4.47$ & $\mathbf{11.21}$ & $\mathbf{84.41}$
    \\\midrule
    \rowcolor{robo_red!7}\multicolumn{14}{l}{\textcolor{robo_red}{\textbf{Fine-Tuning (5\%)}}}
    \\
    \textcolor{gray}{Random} & \textcolor{gray}{$31.22$} & \textcolor{gray}{$77.13$} & \textcolor{gray}{$50.32$} & \textcolor{gray}{$\mathbf{12.36}$} & \textcolor{gray}{$1.26$} & \textcolor{gray}{$0.00$} & \textcolor{gray}{$86.03$} & \textcolor{gray}{$41.22$} & \textcolor{gray}{$21.48$} & \textcolor{gray}{$50.67$} & \textcolor{gray}{$2.96$} & \textcolor{gray}{$0.04$} & \textcolor{gray}{$71.38$}
    \\
    MaskCLIP \cite{zhou2022extract} & $37.03$ & $91.09$ & $60.52$ & $4.35$ & $11.90$ & $\mathbf{11.73}$ & $81.24$ & $23.56$ & $61.77$ & $45.93$ & $2.75$ & $12.45$ & $79.58$
    \\
    FC-CLIP \cite{yu2023fc} & $43.71$ & $92.91$ & $\mathbf{71.21}$ & $10.84$ & $0.00$ & $5.60$ & $\mathbf{90.11}$ & $57.54$ & $\mathbf{71.30}$ & $\mathbf{61.04}$ & $\mathbf{11.41}$ & $8.81$ & $\mathbf{86.38}$
    \\
    \textbf{OpenESS (Ours)} & $\mathbf{44.97}$ & $\mathbf{93.58}$ & $70.18$ & $8.44$ & $\mathbf{18.22}$ & $11.01$ & $89.72$ & $\mathbf{57.76}$ & $67.44$ & $56.06$ & $9.59$ & $\mathbf{12.70}$ & $85.46$
    \\\midrule
    \rowcolor{robo_red!7}\multicolumn{14}{l}{\textcolor{robo_red}{\textbf{Fine-Tuning (10\%)}}}
    \\
    \textcolor{gray}{Random} & \textcolor{gray}{$33.67$} & \textcolor{gray}{$85.79$} & \textcolor{gray}{$49.85$} & \textcolor{gray}{$6.78$} & \textcolor{gray}{$8.00$} & \textcolor{gray}{$\mathbf{15.51}$} & \textcolor{gray}{$80.78$} & \textcolor{gray}{$25.72$} & \textcolor{gray}{$58.18$} & \textcolor{gray}{$29.97$} & \textcolor{gray}{$0.82$} & \textcolor{gray}{$8.93$} & \textcolor{gray}{$76.69$}
    \\
    MaskCLIP \cite{zhou2022extract} & $38.83$ & $92.34$ & $69.96$ & $3.64$ & $5.85$ & $12.98$ & $82.23$ & $23.61$ & $66.39$ & $53.23$ & $3.47$ & $13.46$ & $82.36$
    \\
    FC-CLIP \cite{yu2023fc} & $44.09$ & $93.62$ & $72.86$ & $\mathbf{10.88}$ & $0.00$ & $8.23$ & $\mathbf{89.81}$ & $\mathbf{57.05}$ & $\mathbf{71.95}$ & $60.64$ & $9.58$ & $10.42$ & $86.66$
    \\
    \textbf{OpenESS (Ours)} & $\mathbf{46.25}$ & $\mathbf{93.92}$ & $\mathbf{73.34}$ & $8.13$ & $\mathbf{18.61}$ & $15.41$ & $89.03$ & $52.56$ & $71.76$ & $\mathbf{61.71}$ & $\mathbf{9.99}$ & $\mathbf{14.26}$ & $\mathbf{86.72}$
    \\\midrule
    \rowcolor{robo_red!7}\multicolumn{14}{l}{\textcolor{robo_red}{\textbf{Fine-Tuning (20\%)}}}
    \\
    \textcolor{gray}{Random} & \textcolor{gray}{$41.31$} & \textcolor{gray}{$91.08$} & \textcolor{gray}{$67.90$} & \textcolor{gray}{$4.68$} & \textcolor{gray}{$17.90$} & \textcolor{gray}{$\mathbf{17.41}$} & \textcolor{gray}{$85.11$} & \textcolor{gray}{$43.24$} & \textcolor{gray}{$66.62$} & \textcolor{gray}{$43.95$} & \textcolor{gray}{$5.03$} & \textcolor{gray}{$11.55$} & \textcolor{gray}{$82.99$}
    \\
    MaskCLIP \cite{zhou2022extract} & $42.40$ & $93.19$ & $72.49$ & $5.52$ & $18.21$ & $16.17$ & $84.29$ & $35.04$ & $69.44$ & $54.47$ & $2.43$ & $15.15$ & $84.09$
    \\
    FC-CLIP \cite{yu2023fc} & $47.77$ & $91.05$ & $70.90$ & $7.04$ & $\mathbf{21.10}$ & $14.84$ & $\mathbf{91.13}$ & $\mathbf{64.28}$ & $71.62$ & $61.73$ & $\mathbf{13.25}$ & $\mathbf{18.55}$ & $86.95$
    \\
    \textbf{OpenESS (Ours)} & $\mathbf{48.28}$ & $\mathbf{94.21}$ & $\mathbf{74.66}$ & $\mathbf{10.49}$ & $20.46$ & $16.27$ & $90.15$ & $57.66$ & $\mathbf{73.71}$ & $\mathbf{63.95}$ & $11.20$ & $18.29$ & $\mathbf{87.57}$
    \\\bottomrule
    \end{tabular}}
    \label{tab:per_class_dsec11_efficient_frame2voxel}
    \vspace{0.1cm}
\end{table*}

\subsection{Evaluation Configuration}
Following the convention, we use the Intersection-over-Union (\texttt{IoU}) metric to measure the semantic segmentation performance for each semantic class. The IoU score can be calculated via the following equation:

\begin{equation}
    \texttt{IoU} = \frac{TP}{TP+FP+FN}~,
\end{equation}
where $TP$ (True Positive) denotes pixels correctly classified as belonging to the class; $FP$ (False Positive) denotes pixels incorrectly classified as belonging to the class; and $FN$ (False Negative) denotes pixels that belong to the class but are incorrectly classified as something else.

The \texttt{IoU} metric measures the overlap between the predicted segmentation and the ground truth for a specific class. It returns a value between $0$ (no overlap) and $1$ (perfect overlap). It is a way to summarize the \texttt{mIoU} values for each class into a single metric that captures the overall performance of the model across all classes, \ie, mean IoU (\texttt{mIoU}). The \texttt{mIoU} of a given prediction is calculated as:

\begin{equation}
    \texttt{mIoU} = \frac{1}{C}\sum_{i=1}^C \texttt{IoU}_i~,
\end{equation}
where $C$ is the number of classes and $\texttt{IoU}_i$ denotes the score of class $i$. \texttt{mIoU} provides a balanced measure since each class contributes equally to the final score, regardless of its size or frequency in the dataset. A higher \texttt{mIoU} indicates better semantic segmentation performance. A score of $1$ would indicate perfect segmentation for all classes, while a score of $0$ would imply an absence of correct predictions. In this work, all the compared approaches adopt the same \texttt{mIoU} calculation as in the ESS benchmarks \cite{alonso2019ev-segnet,sun2022ess}. Additionally, we also report the semantic segmentation accuracy (\texttt{Acc}) for the baselines and the proposed framework.

\begin{table*}[t]
    \centering
    \caption{\textbf{The per-class segmentation results} of annotation-efficient event-based semantic segmentation approaches on the test set of \textit{DSEC-Semantic} \cite{sun2022ess}. All approaches adopted the \texttt{frame2recon} representation. Scores reported are IoUs in percentage (\%). For each semantic class under each experimental setting, the best score in each column is highlighted in \textbf{bold}.}
    \vspace{-0.1cm}
    \scalebox{0.865}{\begin{tabular}{r|c|ccccccccccc|c}
    \toprule
    \textbf{Method} & \rotatebox{90}{\textbf{mIoU}} & \rotatebox{90}{background~} & \rotatebox{90}{building} & \rotatebox{90}{fence} & \rotatebox{90}{person} & \rotatebox{90}{pole} & \rotatebox{90}{road} & \rotatebox{90}{sidewalk} & \rotatebox{90}{vegetation} & \rotatebox{90}{car} & \rotatebox{90}{wall} & \rotatebox{90}{traffic-sign} & \textbf{Acc}
    \\\midrule\midrule
    \rowcolor{robo_blue!7}\multicolumn{14}{l}{\textcolor{robo_blue}{\textbf{Linear Probing}}}
    \\
    \textcolor{gray}{Random} & \textcolor{gray}{$6.22$} & \textcolor{gray}{$7.55$} & \textcolor{gray}{$5.48$} & \textcolor{gray}{$0.00$} & \textcolor{gray}{$0.00$} & \textcolor{gray}{$0.00$} & \textcolor{gray}{$39.79$} & \textcolor{gray}{$0.00$} & \textcolor{gray}{$15.64$} & \textcolor{gray}{$0.01$} & \textcolor{gray}{$0.00$} & \textcolor{gray}{$0.00$} & \textcolor{gray}{$36.60$}
    \\
    MaskCLIP \cite{zhou2022extract} & $27.09$ & $59.82$ & $62.14$ & $1.60$ & $0.00$ & $4.54$ & $69.71$ & $5.34$ & $47.85$ & $38.51$ & $0.40$ & $8.12$ & $70.59$
    \\
    FC-CLIP \cite{yu2023fc} & $40.08$ & $\mathbf{89.22}$ & $\mathbf{69.08}$ & $\mathbf{14.62}$ & $\mathbf{26.90}$ & $0.00$ & $83.14$ & $21.79$ & $\mathbf{69.56}$ & $57.78$ & $\mathbf{7.86}$ & $0.92$ & $82.70$
    \\
    \textbf{OpenESS (Ours)} & $\mathbf{44.08}$ & $88.56$ & $61.43$ & $6.05$ & $21.54$ & $\mathbf{12.36}$ & $\mathbf{91.43}$ & $\mathbf{63.04}$ & $64.01$ & $\mathbf{60.52}$ & $6.18$ & $\mathbf{9.76}$ & $\mathbf{84.48}$
    \\\midrule
    \rowcolor{robo_red!7}\multicolumn{14}{l}{\textcolor{robo_red}{\textbf{Fine-Tuning (1\%)}}}
    \\
    \textcolor{gray}{Random} & \textcolor{gray}{$23.95$} & \textcolor{gray}{$76.37$} & \textcolor{gray}{$29.59$} & \textcolor{gray}{$1.73$} & \textcolor{gray}{$0.00$} & \textcolor{gray}{$5.75$} & \textcolor{gray}{$78.12$} & \textcolor{gray}{$9.73$} & \textcolor{gray}{$48.96$} & \textcolor{gray}{$11.56$} & \textcolor{gray}{$0.28$} & \textcolor{gray}{$1.38$} & \textcolor{gray}{$69.20$}
    \\
    MaskCLIP \cite{zhou2022extract} & $30.73$ & $79.25$ & $47.26$ & $0.13$ & $1.17$ & $5.04$ & $78.78$ & $19.72$ & $56.13$ & $43.74$ & $1.13$ & $5.70$ & $74.25$
    \\
    FC-CLIP \cite{yu2023fc} & $38.99$ & $87.75$ & $61.48$ & $3.47$ & $4.60$ & $8.06$ & $88.96$ & $55.12$ & $64.41$ & $47.16$ & $\mathbf{3.61}$ & $4.23$ & $82.90$
    \\
    \textbf{OpenESS (Ours)} & $\mathbf{43.17}$ & $\mathbf{87.85}$ & $\mathbf{66.15}$ & $\mathbf{8.82}$ & $\mathbf{21.52}$ & $\mathbf{12.41}$ & $\mathbf{89.36}$ & $\mathbf{55.35}$ & $\mathbf{72.45}$ & $\mathbf{48.76}$ & $3.40$ & $\mathbf{8.81}$ & $\mathbf{84.56}$
    \\\midrule
    \rowcolor{robo_red!7}\multicolumn{14}{l}{\textcolor{robo_red}{\textbf{Fine-Tuning (5\%)}}}
    \\
    \textcolor{gray}{Random} & \textcolor{gray}{$30.42$} & \textcolor{gray}{$80.25$} & \textcolor{gray}{$38.43$} & \textcolor{gray}{$5.50$} & \textcolor{gray}{$13.45$} & \textcolor{gray}{$9.08$} & \textcolor{gray}{$83.45$} & \textcolor{gray}{$30.88$} & \textcolor{gray}{$51.75$} & \textcolor{gray}{$19.53$} & \textcolor{gray}{$0.16$} & \textcolor{gray}{$2.19$} & \textcolor{gray}{$73.65$}
    \\
    MaskCLIP \cite{zhou2022extract} & $36.33$ & $85.80$ & $60.43$ & $2.60$ & $8.70$ & $7.47$ & $83.10$ & $34.04$ & $64.80$ & $39.60$ & $3.07$ & $\mathbf{10.00}$ & $80.37$
    \\
    FC-CLIP \cite{yu2023fc} & $43.34$ & $88.28$ & $64.90$ & $6.94$ & $\mathbf{20.96}$ & $\mathbf{9.58}$ & $\mathbf{91.18}$ & $\mathbf{62.35}$ & $68.09$ & $52.39$ & $4.93$ & $7.16$ & $84.93$
    \\
    \textbf{OpenESS (Ours)} & $\mathbf{45.58}$ &$\mathbf{89.11}$ & $\mathbf{70.83}$ & $\mathbf{10.92}$ & $20.21$ & $1.99$ & $91.04$ & $60.76$ & $\mathbf{72.07}$ & $\mathbf{67.91}$ & $\mathbf{12.90}$ & $3.69$ & $\mathbf{86.93}$
    \\\midrule
    \rowcolor{robo_red!7}\multicolumn{14}{l}{\textcolor{robo_red}{\textbf{Fine-Tuning (10\%)}}}
    \\
    \textcolor{gray}{Random} & \textcolor{gray}{$34.11$} & \textcolor{gray}{$81.85$} & \textcolor{gray}{$46.28$} & \textcolor{gray}{$4.87$} & \textcolor{gray}{$11.30$} & \textcolor{gray}{$10.20$} & \textcolor{gray}{$85.32$} & \textcolor{gray}{$43.16$} & \textcolor{gray}{$55.34$} & \textcolor{gray}{$32.72$} & \textcolor{gray}{$1.28$} & \textcolor{gray}{$2.90$} & \textcolor{gray}{$77.48$}
    \\
    MaskCLIP \cite{zhou2022extract} & $40.13$ & $87.31$ & $62.54$ & $4.93$ & $5.09$ & $\mathbf{12.86}$ & $88.30$ & $50.60$ & $64.74$ & $55.21$ & $0.32$ & $9.51$ & $83.52$
    \\
    FC-CLIP \cite{yu2023fc} & $45.35$ & $89.71$ & $69.00$ & $6.64$ & $22.37$ & $8.33$ & $91.20$ & $64.09$ & $69.34$ & $61.73$ & $7.23$ & $9.19$ & $86.29$
    \\
    \textbf{OpenESS (Ours)} & $\mathbf{48.94}$ & $\mathbf{90.63}$ & $\mathbf{71.68}$ & $\mathbf{12.41}$ & $\mathbf{29.32}$ & $9.42$ & $\mathbf{92.53}$ & $\mathbf{66.19}$ & $\mathbf{73.76}$ & $\mathbf{69.03}$ & $\mathbf{10.71}$ & $\mathbf{12.71}$ & $\mathbf{87.84}$
    \\\midrule
    \rowcolor{robo_red!7}\multicolumn{14}{l}{\textcolor{robo_red}{\textbf{Fine-Tuning (20\%)}}}
    \\
    \textcolor{gray}{Random} & \textcolor{gray}{$39.25$} & \textcolor{gray}{$87.14$} & \textcolor{gray}{$61.80$} & \textcolor{gray}{$6.77$} & \textcolor{gray}{$3.51$} & \textcolor{gray}{$13.19$} & \textcolor{gray}{$88.53$} & \textcolor{gray}{$56.12$} & \textcolor{gray}{$61.95$} & \textcolor{gray}{$44.65$} & \textcolor{gray}{$1.29$} & \textcolor{gray}{$6.84$} & \textcolor{gray}{$82.51$}
    \\
    MaskCLIP \cite{zhou2022extract} & $43.37$ & $89.83$ & $69.80$ & $7.07$ & $8.93$ & $10.67$ & $88.88$ & $52.65$ & $70.71$ & $60.03$ & $3.10$ & $15.39$ & $85.69$
    \\
    FC-CLIP \cite{yu2023fc} & $47.18$ & $91.20$ & $71.39$ & $\mathbf{11.53}$ & $24.92$ & $9.60$ & $91.58$ & $63.88$ & $71.52$ & $63.44$ & $7.55$ & $12.36$ & $87.07$
    \\
    \textbf{OpenESS (Ours)} & $\mathbf{49.74}$ & $\mathbf{91.28}$ & $\mathbf{73.43}$ & $10.69$ & $\mathbf{27.18}$ & $\mathbf{13.85}$ & $\mathbf{92.84}$ & $\mathbf{67.59}$ & $\mathbf{74.20}$ & $\mathbf{69.22}$ & $\mathbf{10.62}$ & $\mathbf{16.21}$ & $\mathbf{88.26}$
    \\\bottomrule
    \end{tabular}}
    \label{tab:per_class_dsec11_efficient_frame2recon}
    \vspace{0.1cm}
\end{table*}

\section{Additional Experimental Results}
In this section, we provide the class-wise IoU scores for the experiments conducted in the main body of this paper.

\subsection{Annotation-Free ESS}
The per-class zero-shot event-based semantic segmentation results are shown in \cref{tab:per_class_dsec11}. For almost every semantic class, we observe that the proposed OpenESS achieves much higher IoU scores than MaskCLIP \cite{zhou2022extract} and FC-CLIP \cite{yu2023fc}. This validates the effectiveness of OpenESS for conducting efficient and accurate event-based semantic segmentation without using either the event or frame labels.

\subsection{Annotation-Efficient ESS}
The per-class linear probing event-based semantic segmentation results are shown in the first block of \cref{tab:per_class_dsec11_efficient_frame2voxel} and \cref{tab:per_class_dsec11_efficient_frame2recon}. Specifically, compared to the random initialization baseline, a self-supervised pre-trained network always provides better features. The quality of representation learning often determines the linear probing performance. The network pre-trained using our frame-to-event contrastive distillation and text-to-event consistency regularization tends to achieve higher event-based semantic segmentation results than MaskCLIP \cite{zhou2022extract} and FC-CLIP \cite{yu2023fc}. Notably, such improvements are holistic across almost all eleven semantic classes in the dataset. These results validate the effectiveness of the proposed OpenESS framework in tackling the challenging event-based semantic segmentation task.

The per-class annotation-efficient event-based semantic segmentation results of the \texttt{frame2vodel} and \texttt{frame2recon} settings under 1\%, 5\%, 10\%, and 20\% annotation budgets are shown in \cref{tab:per_class_dsec11_efficient_frame2voxel} and \cref{tab:per_class_dsec11_efficient_frame2recon}, respectively. Similar to the findings and conclusions drawn above, we observe clear superiority of the proposed OpenESS framework over the random initialization, MaskCLIP \cite{zhou2022extract}, and FC-CLIP \cite{yu2023fc} approaches. Such consistent performance improvements validate again the effectiveness and superiority of the proposed frame-to-event contrastive distillation and text-to-event consistency regularization. We hope our framework can lay a solid foundation for future works in the established annotation-efficient event-based semantic segmentation.

\section{Qualitative Assessment}
In this section, we provide sufficient qualitative examples to further attest to the effectiveness and superiority of the proposed framework.

\subsection{Open-Vocabulary Examples}
The key advantage of our proposed OpenESS framework is its capability to leverage open-world vocabularies from the CLIP text embedding space. Unlike prior event-based semantic segmentation, which relies on pre-defined and fixed categories, our open-vocabulary segmentation aims to understand and categorize image regions into a broader, potentially unlimited range of categories. We provide more open-vocabulary examples in \cref{fig:qualitative_ov}. As can be observed, given proper text prompts like \textit{``road''}, \textit{``sidewalk''}, and \textit{``building''}, our proposed OpenESS framework is capable of generating semantically meaningful attention maps for depicting the corresponding regions. Such a flexible framework can be further adapted to new or unseen categories without the need for extensive retraining, which is particularly beneficial in dynamic environments where new objects or classes might frequently appear. Additionally, the open-vocabulary segmentation pipeline allows users to work with a more extensive range of objects and concepts, enhancing the user experience and interaction capabilities.

\subsection{Visual Comparisons}
In this section, we provide more qualitative comparisons of our proposed OpenESS framework over prior works \cite{zhou2022extract,sun2022ess} on the DSEC-Semantic dataset. Specifically, the visual comparisons are shown in \cref{fig:qualitative_supp_01} and  \cref{fig:qualitative_supp_02}. As can be observed, OpenESS shows superior event-based semantic segmentation performance over prior works across a wide range of event scenes under different lighting and weather conditions. Such consistent segmentation performance improvements provide a solid foundation to validate the effectiveness and superiority of the proposed frame-to-event contrastive distillation and text-to-event consistency regularization. For additional qualitative comparisons, kindly refer to \cref{sec:demos}.

\subsection{Failure Cases}
\begin{figure*}[t]
    \begin{center}
    \includegraphics[width=1.0\linewidth]{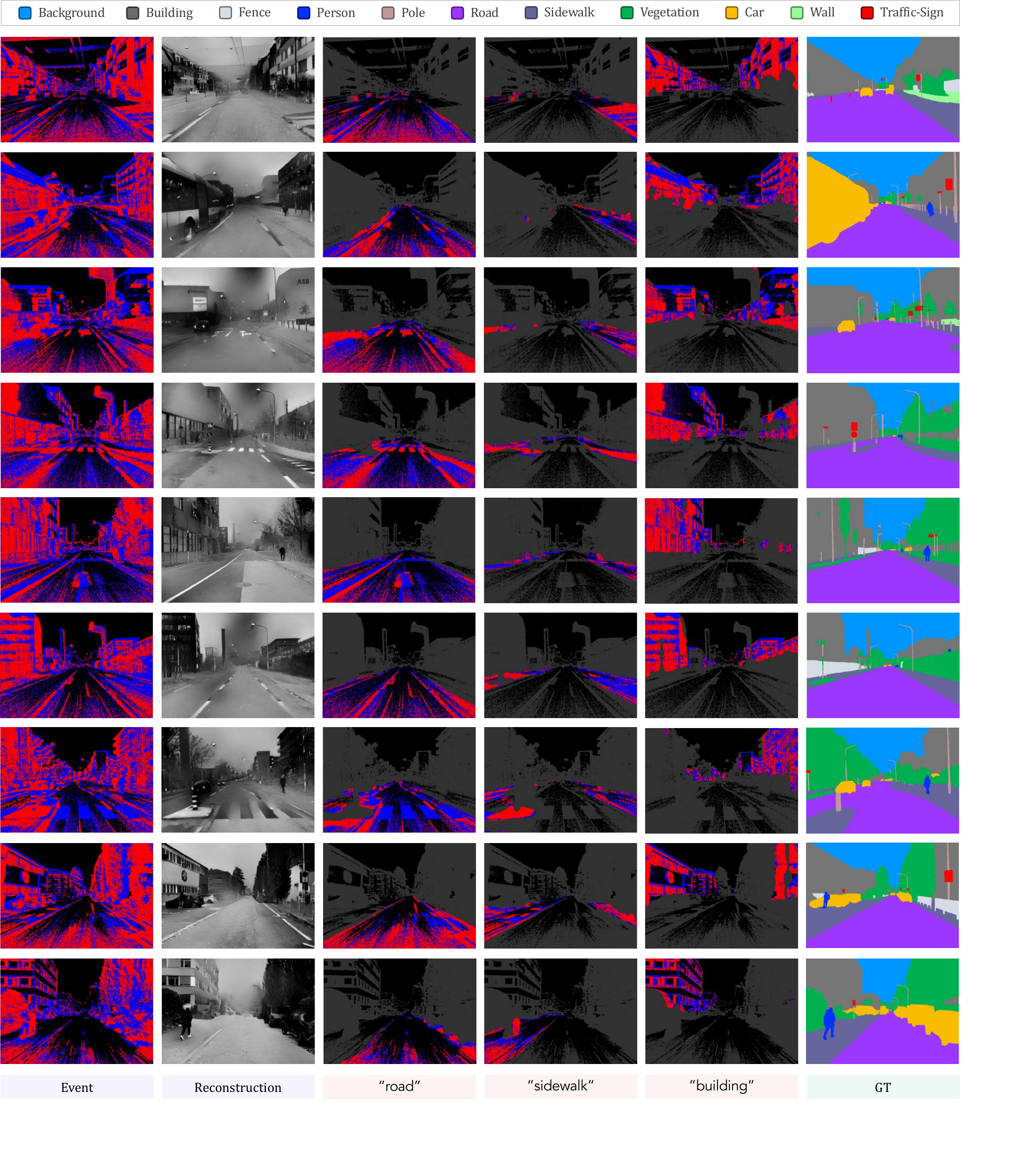}
    \end{center}
    \vspace{-0.4cm}
    \caption{\textbf{Qualitative examples} of the language-guided attention maps generated by the proposed OpenESS framework. For each sample, the regions with a high similarity score to the text prompts are highlighted. Best viewed in colors and zoomed-in for additional details.}
    \label{fig:qualitative_ov}
\end{figure*}

\begin{figure*}[t]
    \begin{center}
    \includegraphics[width=1.0\linewidth]{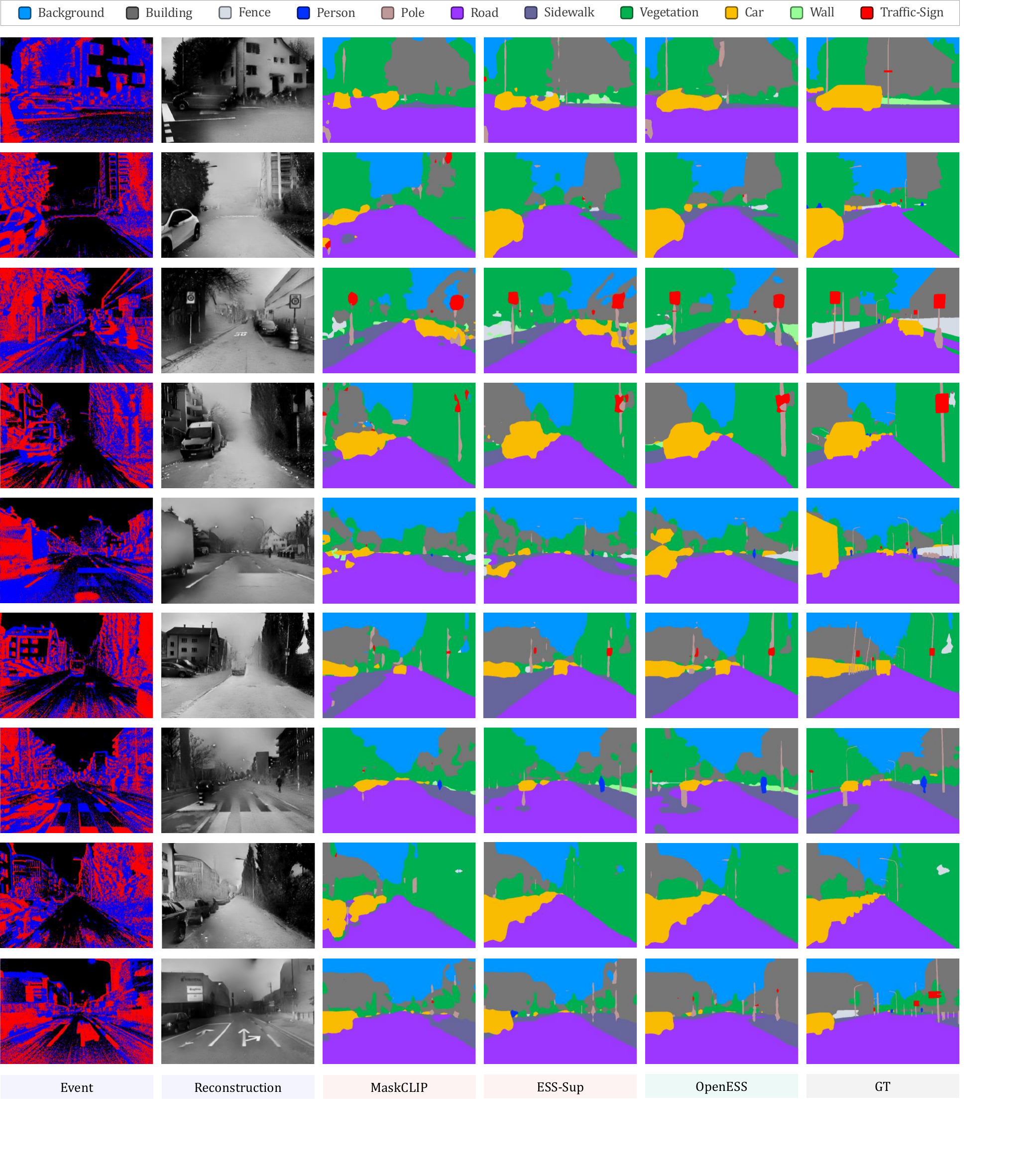}
    \end{center}
    \vspace{-0.4cm}
    \caption{\textbf{Qualitative comparisons} (1/2) among different ESS approaches on the \textit{test} set of \textit{DSEC-Semantic} \cite{sun2022ess}. Best viewed in colors.}
    \label{fig:qualitative_supp_01}
\end{figure*}

\begin{figure*}[t]
    \begin{center}
    \includegraphics[width=1.0\linewidth]{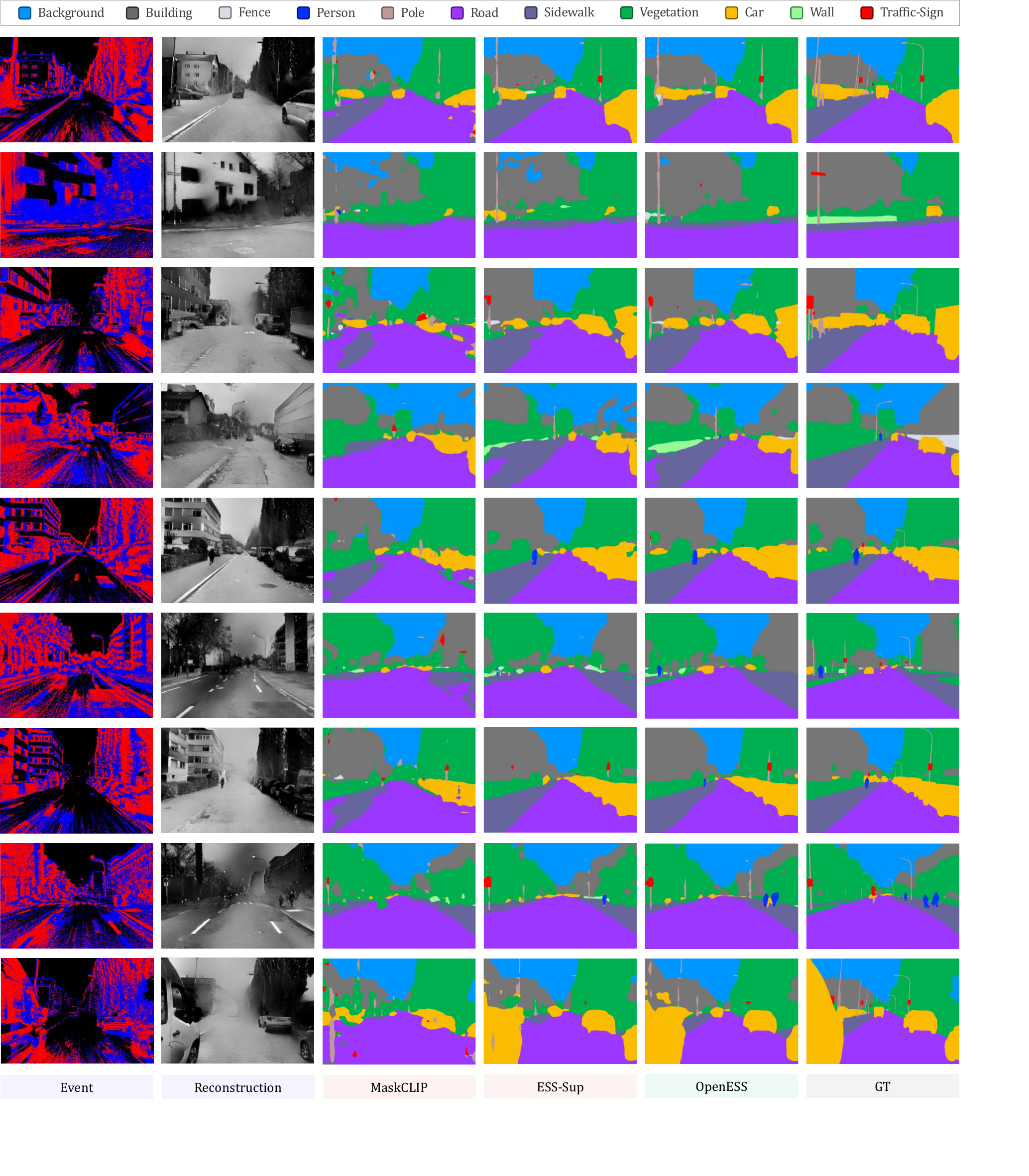}
    \end{center}
    \vspace{-0.4cm}
    \caption{\textbf{Qualitative comparisons} (2/2) among different ESS approaches on the \textit{test} set of \textit{DSEC-Semantic} \cite{sun2022ess}. Best viewed in colors.}
    \label{fig:qualitative_supp_02}
\end{figure*}

As can be observed from \cref{fig:qualitative_ov}, \cref{fig:qualitative_supp_01}, and  \cref{fig:qualitative_supp_02}, the existing event-based semantic segmentation approaches still have room for further improvements. Similar to the conventional semantic segmentation task, it is often hard to accurately segment the boundaries between the semantic objects and backgrounds. In the context of event-based semantic segmentation, such a problem tends to be particularly overt. Unlike traditional cameras that capture dense, synchronous frames, event cameras generate sparse, asynchronous events, which brings extra difficulties for accurate boundary segmentation. Meanwhile, the current framework finds it hard to accurately predict the minor classes, such as \textit{fence}, \textit{pole}, \textit{wall}, and \textit{traffic-sign}. We believe these are potential directions that future works can explore to further improve the event-based semantic segmentation performance on top of existing frameworks.

\subsection{Video Demos}
\label{sec:demos}
In addition to the qualitative examples shown in the main body and this supplementary file, we also provide several video clips to further validate the effectiveness and superiority of the proposed approach. Specifically, we provide three video demos in the attachment, named \texttt{demo1.mp4}, \texttt{demo2.mp4}, and \texttt{demo3.mp4}. The first two video demos show open-vocabulary event-based semantic segmentation examples using the class names and open-world vocabularies as the input text prompts, respectively. The third video demo contains qualitative comparisons of the semantic segmentation predictions among our proposed OpenESS and prior works. All the provided video sequences validate again the unique advantage of the proposed open-vocabulary event-based semantic segmentation framework. Kindly refer to our GitHub repository\footnote{\url{https://github.com/ldkong1205/OpenESS}} for additional details on accessing these video demos.

\section{Broader Impact}
In this section, we elaborate on the positive societal influence and potential limitations of the proposed open-vocabulary event-based semantic segmentation framework.

\subsection{Positive Societal influence}
Event-based cameras can capture extremely fast motions that traditional cameras might miss, making them ideal for dynamic environments. In robotics, this leads to better object detection and scene understanding, enhancing the capabilities of robots in the manufacturing, healthcare, and service industries. In autonomous driving, event-based semantic segmentation provides high temporal resolution and low latency, which is crucial for detecting sudden changes in the environment. This can lead to faster and more accurate responses, potentially reducing accidents and enhancing road safety. Our proposed OpenESS is designed to reduce the annotation budget and training burden of existing event-based semantic segmentation approaches. We believe such an efficient way of learning helps increase the scalability of event-based semantic segmentation systems and in turn contributes positively to impact society by enhancing safety, efficiency, and performance in various aspects.

\subsection{Potential Limitation}
Although our proposed framework is capable of conducting annotation-free and open-vocabulary event-based semantic segmentation and achieves promising performance, there tend to exist several potential limitations. Firstly, our current framework requires the existence of synchronized event and RGB cameras, which might not be maintained by some older event camera systems. Secondly, we directly adopt the standard text prompt templates to generate the text embedding, where a more sophisticated design could further improve the open-vocabulary learning ability of the existing framework. Thirdly, there might still be some self-conflict problems in our frame-to-event contrastive distillation and text-to-event consistency regularization. The design of a better representation learning paradigm on the event-based data could further resolve these issues. We believe these are promising directions that future works can explore to further improve the current framework.

\section{Public Resources Used}
\label{sec:public-resources-used}

In this section, we acknowledge the use of public resources, during the course of this work.

\subsection{Public Datasets Used}
We acknowledge the use of the following public datasets, during the course of this work:

\begin{itemize}
    \item DSEC\footnote{\url{https://dsec.ifi.uzh.ch}}\dotfill CC BY-SA 4.0
    \item DSEC-Semantic\footnote{\url{https://dsec.ifi.uzh.ch/dsec-semantic}}\dotfill CC BY-SA 4.0
    \item DDD17\footnote{\url{http://sensors.ini.uzh.ch/news_page/DDD17.html}}\dotfill CC BY-SA 4.0
    \item DDD17-Seg\footnote{\url{https://github.com/Shathe/Ev-SegNet}}\dotfill Unknown
    \item E2VID-Driving\footnote{\url{https://rpg.ifi.uzh.ch/E2VID.html}}\dotfill GNU General Public License v3.0
\end{itemize}

\subsection{Public Implementations Used}
We acknowledge the use of the following public implementations, during the course of this work:

\begin{itemize}
    \item ESS\footnote{\url{https://github.com/uzh-rpg/ess}}\dotfill GNU General Public License v3.0
    \item E2VID\footnote{\url{https://github.com/uzh-rpg/rpg_e2vid}}\dotfill GNU General Public License v3.0
    \item HMNet\footnote{\url{https://github.com/hamarh/HMNet_pth}}\dotfill BSD 3-Clause License
    \item EV-SegNet\footnote{\url{https://github.com/Shathe/Ev-SegNet}}\dotfill Unknown
    \item SNN-Segmentation\footnote{\url{https://github.com/Intelligent-Computing-Lab-Yale/SNN-Segmentation}}\dotfill Unknown
    \item CLIP\footnote{\url{https://github.com/openai/CLIP}}\dotfill MIT License
    \item MaskCLIP\footnote{\url{https://github.com/chongzhou96/MaskCLIP}}\dotfill Apache License 2.0
    \item FC-CLIP\footnote{\url{https://github.com/bytedance/fc-clip}}\dotfill Apache License 2.0
    \item SLIC-Superpixels\footnote{\url{https://github.com/PSMM/SLIC-Superpixels}}\dotfill Unknown
    \item Segment-Anything\footnote{\url{https://github.com/facebookresearch/segment-anything}}\dotfill Apache License 2.0
\end{itemize}

\clearpage

\clearpage

\clearpage

\clearpage
{\small
 \bibliographystyle{ieeenat_fullname}
 \bibliography{main}
}

\end{document}